\documentclass[11pt]{article}

\usepackage[preprint]{acl}

\usepackage{times}
\usepackage{latexsym}

\usepackage[T1]{fontenc}

\usepackage[utf8]{inputenc}

\usepackage{microtype}

\usepackage{inconsolata}

\usepackage{graphicx}
\usepackage{enumitem}
\usepackage{amsmath,amssymb,mathtools}
\usepackage{booktabs,tabularx,longtable,array,multirow}
\usepackage{enumitem}
\usepackage{xcolor}
\usepackage{hyperref}
\usepackage{microtype}
\usepackage{makecell}
\usepackage{graphicx}
\usepackage{subcaption}
\usepackage{listings}
\usepackage[most]{tcolorbox}
\makeatletter
\@ifpackageloaded{lineno}{%
  \@ifpackagelater{lineno}{2023/05/19}{}{%
    \def\@LN@depthbox{%
      \ifdim\@tempdima=-1000pt
      \else
        \dp\@tempboxa=\@tempdima
        \nointerlineskip
        \kern-\@tempdima
      \fi
      \box\@tempboxa}%
  }%
}{}
\makeatother
\lstdefinestyle{prompt}{basicstyle=\ttfamily\scriptsize,breaklines=true,columns=fullflexible,keepspaces=true,upquote=true,breakindent=1em,aboveskip=0pt,belowskip=0pt,literate={—}{{\textemdash}}1}
\newtcblisting{promptbox}[1]{enhanced,breakable,listing only,listing options={style=prompt},title={#1},fonttitle=\bfseries\footnotesize,colback=white,colframe=black!75,colbacktitle=black!75,coltitle=white,boxrule=0.5pt,arc=1pt,left=3pt,right=3pt,top=2pt,bottom=2pt,toptitle=1pt,bottomtitle=1pt,before skip=6pt,after skip=6pt}
\usepackage{graphicx}

\newcommand{\method}{CVD}

\title{Learning When Human Decisions Matter: Executor-Relative Consultation Value for LLM Agents}

\author{Shuli Jiang, Khushbu Pahwa, Yuting Zhang,\\
\textbf{Yi Zhang, Wei Xia, Stefano Soatto} \\
  AWS Agentic AI \\
  \texttt{\{shulij, pahwakhu, yutingzh, yizhngn, wxia, soattos\}@amazon.com}}

\begin{document}
\maketitle

\begin{abstract}
    Large language models increasingly automate complex tasks, but human decision capacity does not scale with their capabilities. In many applications, humans remain essential for providing private context, authority, or judgments unavailable to the agent.
    We formulate \textit{human decision capacity allocation}: given multiple consultation opportunities and a limited budget, at which decisions should an LLM agent consult a human?
    The central challenge is estimating \textit{consultation value}, which is executor-dependent because the same human input may benefit different downstream agents differently. 
    We introduce \emph{Consultation Value Distillation (\method{})}, which estimates consultation values offline using Monte Carlo executor rollouts and distills these values into lightweight predictors for budgeted allocation, requiring no rollout-based value estimation at deployment. 
    Across five batch and three sequential settings with two frontier executors, \method{} improves allocation through joint and continuation-aware value learning, while independent consultation values remain competitive in several settings. 
    \method{} also broadly outperforms direct prompting and Value of Information \cite{dong2026voi} where applicable, remains robust with substantially less rollout supervision, and enables nearly lossless approximate allocation at low search cost.
\end{abstract}

\section{Introduction}
\label{sec:introduction}

\begin{figure*}
    \centering
    \includegraphics[width=\linewidth]{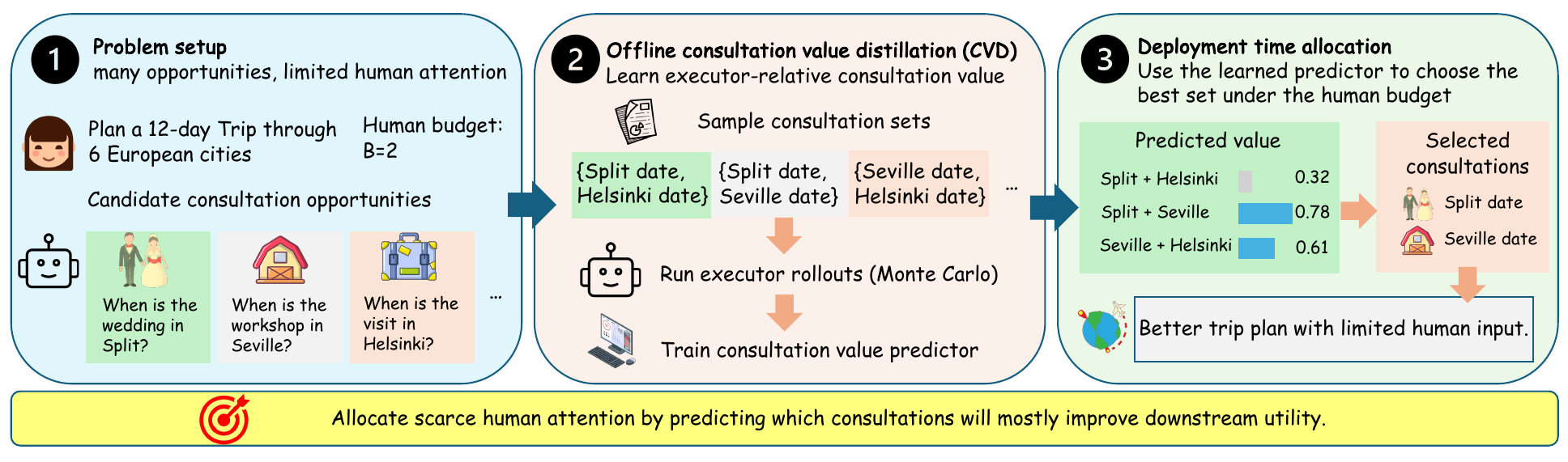}
    \vspace{-20pt}
    \caption{\textbf{Consultation Value Distillation (\method{}) for human decision-capacity allocation.} Given multiple consultation opportunities and a limited human budget, \method{} learns from offline executor rollouts which human inputs most improve downstream utility, then uses the distilled value predictor to allocate consultation at deployment without rollout-based value estimation.}
    \vspace{-10pt}
    \label{fig:banner}
\end{figure*}

Large language model (LLM) agents are increasingly capable of performing complex knowledge-work tasks, including repository-level software engineering, natural-language planning, and collaborative document creation \citep{jimenez2024swebench,yang2024sweagent,zheng2024naturalplan,wu2025collabllm}. As the amount of work that agents can generate and execute grows, however, the human capacity available to supervise them does not scale accordingly. Humans remain essential in many real-world workflows because they possess information unavailable to the agent, exercise organizational authority, and provide judgments that cannot be inferred reliably from public context. A software engineer may know what a customer said minutes ago; a manager may know an unreleased roadmap decision; and among several technically valid code changes, only the responsible team may know which trade-off the organization prefers. 
Yet humans cannot be consulted at every decision point surfaced by increasingly productive agents.

Human decision capacity therefore becomes a scarce resource. Importantly, not every decision point deserves the same amount of human attention. Some can be handled reliably by the agent, while others have substantial downstream consequences if the agent acts without missing human information. Moreover, the apparent importance of a decision point is not the same as the value of consulting a human. 
A salient decision may require little human input when the executor already handles it correctly, whereas a routine decision may benefit greatly from a private fact the executor lacks.
The allocation problem is therefore a decision made by the \emph{agent} to decide where it should spend limited human capacity, given multiple consultation opportunities to ask for human input.
The relevant question is not simply \emph{which decisions are important}, but:
\noindent\fbox{%
\parbox{\linewidth}{%
\textit{Given a limited human consultation budget, at which decision points should the agent consult a human?}
}%
}

We study this problem as \textbf{human decision-capacity allocation}. We assume that candidate consultation points have already been identified and focus on allocating a limited human budget among them to obtain the corresponding information, judgment, or authority. The central challenge is estimating \emph{consultation value}, the expected utility gain from resolving a decision or a set of decisions, before consulting. Once these values are known, budgeted allocation is comparatively direct. We observe that consultation value is \emph{executor-relative}: it depends not only on the decision itself, but also on what the executor already knows and how effectively it can use the human input. We study both batch settings, where multiple consultations are chosen jointly and may be complementary or redundant, and sequential settings, where consultation opportunities arrive over time and using capacity now incurs an opportunity cost for future decisions.

Prior work studies selective human intervention and clarification \citep{mozannar2020consistent,hoque2022thriftydagger,zhang2024clamber}.
Most closely, \cite{dong2026voi} formulate multi-turn clarification as a clarify-or-commit process, selecting among candidate questions by their inference-time Value of Information (VoI) relative to communication cost. We allocate a shared consultation budget across multiple decision opportunities, accounting for set-level interactions and sequential opportunity costs. More broadly, our formulation extends resource-aware agentic inference \citep{achille2025agents} from allocating computation to allocating scarce human consultation by downstream value.

To estimate consultation value, we introduce \textbf{Consultation Value Distillation (\method{})}, as illustrated in Figure~\ref{fig:banner}. We sample consultation choices offline and use Monte Carlo executor rollouts to estimate downstream utility, then distill these outcomes into a lightweight predictor using information available at decision time. In batch settings, a permutation-invariant set-value model scores complete consultation sets, capturing complementarity and redundancy among human inputs. In sequential settings, a continuation-aware predictor accounts for both the current consultation and the remaining human budget. At deployment, the allocator uses these predicted values to select consultations under the available budget without rollout-based value estimation. We use a small ensemble of predictors for robustness.

We evaluate \method{} across five batch and three sequential settings spanning six tasks, covering information acquisition, planning, scheduling, preference elicitation, and controlled decision interactions. Main results are replicated under Sonnet~5 and GPT-6.1~Sol, using executor outcomes independent of training supervision. \method{} provides clear gains from joint batch and continuation-aware sequential allocation. At the largest evaluated budgets, it significantly outperforms the Local Value baseline, reaching normalized gain $0.58/0.40$ vs.\ $0.25/0.09$ on batch Animals and $0.91$--$0.96$ vs.\ $0.80$--$0.87$ on sequential ThinkingBox. Local Value remains competitive on several other tasks. At these budgets, our VoI baseline achieves at most $0.18$ normalized gain, well below \method{} on batch and sequential Animals.
Ablations support direct total-value regression as a strong default, while the benefits of explicit decomposition and ranking objectives depend on the task and executor. Ensembling improves robustness.

Finally, we study \method{}'s practicality and reuse under reduced supervision,
approximate deployment-time search, and cross-executor transfer. Under Sonnet~5 at the largest budgets, \method{} retains at least $90\%$ of the default supervision gain using only $62\%/10\%$ of default training calls on
Animals/Trip, and at least $93\%$ using labels for just $5\%$ of Meeting Planning's feasible sets. A greedy allocator with single-swap local search recovers the exact allocation on $90$--$100\%$ of test tasks. Cross-executor transfer is effective when
executors induce similar consultation-value structure, but executor-specific
learning remains important when they use human information differently.

\textbf{Our contributions are:}
\begin{enumerate}[leftmargin=*, itemsep=0mm, topsep=0mm]
\item \textbf{Scarce human decision-capacity allocation.}
We formulate how an LLM agent should decide when to consult humans across multiple candidate decision points under a limited budget of human information, judgment, or authority.

\item \textbf{Consultation Value Distillation (\method{}).} We investigate executor-relative consultation value from offline Monte Carlo executor rollouts and distill it into lightweight batch and sequential allocators that capture interactions among consultation opportunities without rollout-based value estimation at deployment.

\item \textbf{A diverse evaluation suite.}
We evaluate \method{} across five batch and three sequential settings spanning
six tasks and two strong executors, Sonnet~5 and GPT-6.1~Sol.

\item \textbf{A comprehensive characterization of learning, efficiency, and transfer.}
We study alternative approaches to learning consultation value and quantify how \method{} scales with fewer executor rollouts, fewer labeled consultation sets, approximate subset search, and cross-executor transfer.
\end{enumerate}

\section{Problem Formulation}
\label{sec:problem}

We study how an agent should allocate limited human decision capacity across multiple consultation opportunities. Let $x$ denote the agent-visible task context, $\pi$ a fixed downstream executor, and $U_\pi$ its terminal task utility. Human consultations may reveal private information, judgment, or authority unavailable to the agent.

A task contains $n$ candidate consultation opportunities and may unfold over $T$ decision stages. 
At stage $t$, the allocator observes the interaction history $h_t$, available consultations $C_t$, and remaining budget $b_t$, and selects $S_t\subseteq C_t$ with $|S_t|\leq b_t$. The corresponding human responses for decisions in $S_t$ are then revealed to the executor.

For a feasible $S_t$, let $Q^*_{\pi,t}(S_t\mid h_t,b_t)$ denote the expected terminal utility from choosing $S_t$ now and allocating the remaining budget optimally thereafter. 
The expectation marginalizes over human-private information and responses, executor stochasticity, and environment transitions. 
We define its \textit{consultation value} relative to skipping the current opportunity as
\begin{align}
    &F_{\pi,t}(S_t \mid h_t, b_t)\\
    \nonumber
    &= Q^*_{\pi,t}(S_t \mid h_t, b_t) - Q^*_{\pi, t}(\emptyset \mid h_t, b_t).
\end{align}
We model consultation value as \emph{executor-relative} because the downstream benefit of human input depends on the executor's ability to use it. The same response can therefore have different value for different executors.
See Appendix~\ref{appendix:formal_problem} for a formal recursive formulation.

We study two special cases.  In the \textbf{batch setting}, all $n$ consultation opportunities are available simultaneously, with budget $B$. Let $Q_\pi^{\mathrm{batch}}(S\mid x)=\mathbb{E}[U_\pi\mid x,S]$ denote the expected terminal utility after revealing the consultations in $S$. The consultation value and allocation problem are
\begin{align*}
    &F^{\mathrm{batch}}_\pi(S \mid x)
    = Q^{\mathrm{batch}}_\pi(S \mid x)
- Q^{\mathrm{batch}}_\pi(\emptyset \mid x), \\
    &S^*=
\arg\max_{|S|\le B}
F_\pi^{\mathrm{batch}}(S\mid x).
\end{align*}

Because $F_\pi^{\mathrm{batch}}$ assigns value to the selected set as a whole, it can capture complementarity or redundancy among consultations.

In the \textbf{sequential setting}, the allocator selects at most one consultation from $C_t$ per stage, and $\emptyset$ denotes no consultation at that stage. 
Let $Q_{\pi,t}^{\mathrm{seq}}$ denote the corresponding specialization of $Q^*_{\pi,t}$. 
For a candidate $q_t\in C_t$, its continuation-aware consultation value is thus
\begin{align*}
     &F^{\text{seq}}_{\pi,t}(\{q_t\} \mid h_t, b_t) \\
     \nonumber
     &= Q_{\pi,t}^{\text{seq}}(\{q_t\} \mid h_t, b_t)
- Q_{\pi,t}^{\text{seq}}(\emptyset \mid h_t, b_t).
\end{align*}
Accounting for future budget use, this value captures the benefit of consulting now and the opportunity cost of human capacity that may be more useful later.
When $b_t>0$, the optimal allocator selects the candidate with the highest positive value, or the empty action if none is positive.

\section{Consultation Value Distillation (\method)}
\label{sec:method}

Consultation value is difficult to infer directly from task context alone: an LLM may recognize that information is missing without knowing whether obtaining it will actually improve its own downstream execution. 
We therefore learn consultation value from observed executor outcomes rather than relying only on direct prompting.
Our approach uses \emph{offline distillation}: for a given executor and task, we collect Monte Carlo rollouts under alternative consultation choices, aggregate them into consultation-value supervision, and distill this supervision into a lightweight predictor. This amortizes the cost of estimating downstream value across many future decisions and avoids repeatedly performing expensive rollout-based evaluation during allocation.

\subsection{Offline Monte Carlo Supervision}
\label{subsec:supervision}

In \textbf{the batch setting},
for each sampled consultation set $S$, we estimate
expected utility using $R_{\rm train}$ independent Monte Carlo samples:
\begin{align*}
    &\widehat{Q}^{\mathrm{batch}}_{\pi}(S \mid x) = \frac{1}{R_{\mathrm{train}}}\sum_{r=1}^{R_{\mathrm{train}}} U_{\pi}^{(r)}(x, S),\\
    &\widehat{F}^{\mathrm{batch}}_\pi(S \mid x) =
\widehat{Q}^{\mathrm{batch}}_\pi(S \mid x)
- \widehat{Q}^{\mathrm{batch}}_\pi (\emptyset \mid x).
\end{align*}
where $U_\pi^{(r)}(x,S)$ is the sample-level estimate of terminal utility. A Monte Carlo sample may aggregate multiple executor calls when marginalizing over discrete hidden-response configurations.
We do not require labels for all possible subsets. For each training task, we may label only a sampled collection of consultation sets. See Section~\ref{sec:practicality} for studies on how performance changes with fewer labeled sets and fewer MC samples per label.

In \textbf{the sequential setting}, for each sampled state with current consultation opportunity $q_t$, we analogously estimate the continuation utility of consulting versus taking the empty action using $R_{\mathrm{train}}$ executor rollouts. Their difference defines the continuation-aware value:
\begin{align*}
    &\widehat{F}^{\mathrm{seq}}_{\pi,t}(\{q_t\} \mid h_t, b_t)\\
    \nonumber
    &= \widehat{Q}^{\mathrm{seq}}_{\pi, t}(\{q_t\} \mid h_t, b_t)
    - \widehat{Q}^{\mathrm{seq}}_{\pi, t}(\emptyset \mid h_t, b_t).
\end{align*}
Unlike a myopic estimate of the immediate usefulness of the current response, the sequential target measures whether spending one unit of human capacity now improves final utility after accounting for future consultation opportunities.

\subsection{Consultation Value Regression}
\label{subsec:regression}

The Monte Carlo estimates above are expensive to obtain but can be generated offline. \method{} distills them into predictors that use only information available before human consultation.

For \textbf{batch allocation}, we train a permutation-invariant set-value model $D_\theta(x,S)$ based on DeepSets \cite{zaheer2017deepsets} to predict $\widehat{F}_\pi^{\mathrm{batch}}(S\mid x)$, using 
squared error regression:
\begin{align*}
    \mathcal{L}^{\mathrm{batch}}(\theta) = \mathbb{E}_{x,S}\!\bigl[
        \bigl(D_{\theta}(x,S) - \widehat{F}^{\mathrm{batch}}_{\pi}(S \mid x)\bigr)^2
    \bigr].
\end{align*}
Predicting the value of the complete set allows \method{} to capture complementarity and redundancy among consultations without explicitly decomposing value into singleton and interaction terms.

For \textbf{sequential allocation}, we analogously distill the continuation estimates into a lightweight continuation-aware consultation value
$G_\phi(h_t,q_t,b_t)$. Depending on the benchmark, this score is learned directly or as the difference between separately supervised consult and skip heads (Appendix~\ref{subsec:model_training}). 
Because the supervision accounts for subsequent use of the remaining budget, $G_\phi$ captures whether consulting now is more valuable than preserving capacity for later opportunities.

To reduce noise from stochastic executor rollouts, we average predictions from an ensemble of five independently initialized models. See Section~\ref{sec:alt_learning} for comparison of this direct value-distillation objective with structured and uncertainty-aware alternative learning approaches.

\subsection{Budgeted Allocation}
\label{subsec:budgeted_allocation}

At deployment, \method{} replaces rollout-based value estimation with the distilled predictor. In batch tasks, it selects the feasible set with highest predicted value. We use exact enumeration in the main experiments and study approximate search in Section~\ref{sec:practicality}. 
In sequential tasks, it selects the consultation in $C_t$ with the highest predicted continuation-aware value if budget remains and this value is positive. Otherwise it takes the empty action.

\section{Benchmarks and Evaluation}
\label{sec:benchmarks_and_evaluation}

\begin{table}[t]
    \centering
    \small
    \setlength{\tabcolsep}{3pt}
    \begin{tabular}{llrr}
        \toprule
        \textbf{Benchmark} &
        \textbf{Regime} &
        \textbf{Train} &
        \textbf{Test} \\
        \midrule
        Animals & Batch, Seq. & 245 & 75 \\
        NaturalPlan Trip Planning & Batch & 521 & 79 \\
        NaturalPlan Meeting Planning & Batch & 80 & 20 \\
        Flight Recommendation & Batch & 113 & 30 \\
        \midrule
        ThinkingBox-D (TB-D) & Batch, Seq. & 60 & 60 \\
        ThinkingBox-H (TB-H) & Seq. & 114 & 30 \\
        \bottomrule
    \end{tabular}
    \caption{
    Benchmark suite. Train counts denote non-test tasks used for allocator development and final training. Test splits are evaluated after methods and hyperparameters are frozen. Animals and ThinkingBox-D share the same splits and executor rollouts across batch and sequential settings.
}
    \vspace{-10pt}
    \label{tab:benchmark_suite}
\end{table}

\subsection{Benchmark Suite}
\label{subsec:benchmark_suite}

We evaluate human decision-capacity allocation across four adapted tasks and two controlled environments as follows, spanning information acquisition, planning, preference elicitation, and scheduling. 
Table~\ref{tab:benchmark_suite} presents a summary of the benchmarks. 
Detailed benchmark construction, candidate-generation procedures, utility definitions and features used by the predictor are provided in Appendix~\ref{appendix:benchmark}. Prompt templates and executor-prompt details are in Appendix~\ref{appendix:prompts}.

\begin{enumerate}[leftmargin=*, topsep=0mm, itemsep=0mm]
    \item \textbf{Animals.} We adapt the Mixed 20Q Animals setting from \cite{dong2026voi},
    where the executor must identify a hidden animal from a pool of candidates and consultations reveal answers to candidate questions.

    \item \textbf{Trip Planning.} We adapt NaturalPlan Trip Planning \citep{zheng2024naturalplan}, where consultations reveal private constraints needed to construct a valid itinerary.

    \item \textbf{Meeting Planning.} We adapt NaturalPlan Meeting Planning \citep{zheng2024naturalplan}, where consultations reveal participant availability for scheduling meetings.

    \item \textbf{Flight Recommendation.} We adapt the setting of \cite{dong2026voi}, based on FLIGHTPREF \citep{lin2022inferringrewards}, where consultations reveal user preferences before the executor recommends a flight.

    \item \textbf{ThinkingBox.} We build controlled allocation environments from ThinkingBox \citep{li2026thinkingbox}, where consultations reveal task-relevant private user information needed by the executor to complete a stateful workflow. TB-D is evaluated in both batch and sequential regimes, while TB-H presents consultation opportunities sequentially, requiring the allocator to decide whether to use limited human capacity now or preserve it for later stages.

\end{enumerate}

\subsection{Baselines}
\label{subsec:baselines}

We compare against three deployable baselines:
\begin{enumerate}[leftmargin=*, topsep=0mm, itemsep=0mm]
    \item \textbf{Prompt} directly asks the LLM to rank or select consultations based on their expected usefulness.

    \item \textbf{Value of Information (VoI)} \citep{dong2026voi}, a model-based approach to explicitly reason over possible human responses and downstream utility. Its inference-time estimator requires a tractable response and task model, making it most directly applicable to benchmarks with structured human responses. 
    We evaluate its released value estimator on Animals and Flight, adapting consultation selection to the shared candidate pool and budget (Appendix~\ref{subsec:model_training}).

    \item \textbf{Local Value} tests whether independently learned consultation values suffice without modeling set-level or continuation-aware value. It uses the same offline MC distillation as \method{}, but predicts executor-relative singleton values independently. Batch allocation selects the highest-valued candidates; sequential allocation uses myopic values with thresholds tuned on non-test tasks.
    
\end{enumerate}

\subsection{Metrics and Evaluation Setup}
\label{subsec:metrics}

Training labels are estimated from $R_{\mathrm{train}}$ executor rollouts, and evaluation uses independent executor outcomes. At deployment, the allocator requires no rollouts. Rollout counts and label reliability are reported in Appendix~\ref{appendix:benchmark_stats}.

Because benchmark utilities use different scales, we report \emph{Normalized Gain} (NG), 
\begin{align*}
    \mathrm{NG}=(\bar U_{\rm method}-\bar U_{\rm random})/(\bar U_{\rm oracle}-\bar U_{\rm random}),
\end{align*}
where $\bar U$ is mean test utility. For batch benchmarks and sequential Animals, random denotes uniform size-$B$ allocation, and oracle is selected on label-half outcomes and scored on independent outcomes. The sequential Animals oracle is adaptive. For sequential TB-D and TB-H, random denotes no consultation, and oracle is optimized over the evaluation completions, giving an empirical adaptive upper bound (Appendix~\ref{appendix:ref_eval}).
For reported results, Prompt and VoI are averaged over three independent stochastic test runs, while Local Value and \method{} are averaged over three independent training runs, each trained only on non-test data and evaluated once on the frozen test set. Thus, all learned allocators are evaluated on test outcomes independent of their training supervision. We clip NG to $[0,1]$ for presentation.
Batch \method{} allows $|S|\leq B$, while Local Value, Prompt, and VoI use
exactly $B$. This freedom is unused outside Flight and hurts \method{} on
Flight (Appendix~\ref{appendix:cardinality}).
Additional details on evaluation,
model specifications, and training configurations are provided in
Appendix~\ref{appendix:metrics_eval}.

\section{Main Results}
\label{sec:main_results}
We evaluate \method\ under two executors: Sonnet~5 (Sonnet) and GPT-6.1~Sol (Sol), across five batch and three sequential settings. Table~\ref{tab:main_results} reports NG on the held-out test set at each benchmark's largest evaluated human decision budget $B$, along with the number of candidate decision points $n$. Results across all budgets, raw utilities and confidence intervals are provided in Appendix~\ref{appendix:full_results}.

\begin{table*}[t]
    \centering
    \small
    \setlength{\tabcolsep}{4.5pt}
    \begin{tabular}{llcc cccc cccc}
        \toprule
        & & & & \multicolumn{4}{c}{\textbf{Sonnet 5}} & \multicolumn{4}{c}{\textbf{GPT-6.1 Sol}} \\
        \cmidrule(lr){5-8}\cmidrule(lr){9-12}
        & & & & & & \multicolumn{2}{c}{\textit{Ours}} & & & \multicolumn{2}{c}{\textit{Ours}} \\
        \cmidrule(lr){7-8}\cmidrule(lr){11-12}
        \textbf{Setting} & \textbf{Benchmark} & $n$ & $B$ & Prompt & VoI & Local Value & \method{} & Prompt & VoI & Local Value & \method{} \\
        \midrule
        \multirow{5}{*}{Batch} & Animals & 6 & 3 & 0.16 & 0.05 & 0.25 & \textbf{0.58}$^{\ast}$ & 0.19 & 0.17 & 0.09 & \textbf{0.40}$^{\ast}$ \\
        & Trip Planning & 3 & 2 & 0.09 & --- & 0.79 & \textbf{0.83} & 0.31 & --- & 0.75 & \textbf{0.80} \\
        & Meeting Planning & 10 & 4 & 0.00 & --- & 0.91 & \textbf{1.00} & 0.22 & --- & \textbf{0.88} & 0.82 \\
        & Flight & 5 & 3 & 0.24 & 0.18 & 0.00 & \textbf{0.33} & \textbf{0.29} & 0.14 & 0.09 & 0.00 \\
        & TB-D & 5 & 1 & 0.58 & --- & \textbf{1.00} & \textbf{1.00} & 0.54 & --- & \textbf{1.00} & 0.97 \\
        \midrule
        \multirow{3}{*}{Sequential} & Animals & 6 & 3 & 0.33 & 0.09 & \textbf{0.42} & 0.40 & 0.19 & 0.13 & \textbf{0.42} & 0.36 \\
        & TB-D & 5 & 2 & 0.75 & --- & 0.87 & \textbf{0.95}$^{\ast}$ & 0.77 & --- & 0.87 & \textbf{0.96}$^{\ast}$ \\
        & TB-H & 9--10 & 2 & 0.51 & --- & 0.80 & \textbf{0.91}$^{\ast}$ & 0.54 & --- & 0.80 & \textbf{0.91}$^{\ast}$ \\
        \bottomrule
    \end{tabular}
    \caption{
    Test normalized gain (clipped to $[0,1]$) at each benchmark's largest evaluated budget $B$ under $n$ candidate consultation opportunities, averaged over 3 runs. Local Value is our singleton value control and \method{} our proposed allocator. Bold denotes the best allocator per executor. 
    --- means VoI's released implementation does not apply to the benchmark. 
    $^{\ast}$ denotes \method{} significantly better than Local Value (paired $95\%$ bootstrap). Full results are in Appendix~\ref{appendix:full_results}.
    }
    \vspace{-10pt}
    \label{tab:main_results}
\end{table*}

\paragraph{Simpler local values suffice when interactions are weak or exposed by feedback.}
When only one consultation is selected, as in batch TB-D, \method{} and Local Value are essentially equivalent and recover nearly all available gain. On Trip and Meeting Planning, \method{} remains competitive but its advantage is not statistically significant. Sequential Animals provides the complementary case: Local Value slightly outperforms \method{} ($0.42$ versus $0.40/0.36$, not significant), suggesting that conditioning each decision on previously observed answers already captures much of the interaction among questions.

\paragraph{Comparison with prompting and VoI.}
Direct prompting is substantially weaker than learned allocation on most benchmarks, suggesting that task knowledge alone does not identify the most useful consultations.
On batch Animals, \method{} substantially outperforms the VoI baseline under both executors ($0.58$ vs.\ $0.05$ and $0.40$ vs.\ $0.17$), with a similar gap in the sequential setting.
On Flight under Sonnet, \method{} also performs best ($0.33$ vs.\ $0.24$ Prompt and $0.18$ VoI).
Allowing VoI to generate follow-up questions beyond our predefined candidates improves its Sonnet~5 performance on sequential Animals (Appendix~\ref{subsec:model_training}).

\paragraph{The main exception.}
\method{} fails on Flight Recommendation under Sol, obtaining clipped NG $0$
versus $0.29$ for prompting and $0.09$ for Local Value, although neither
difference is significant. Unlike on other batch benchmarks, \method{} often
stops before exhausting the budget; forcing exactly $B$ consultations improves
performance (Appendix~\ref{appendix:cardinality}). Sol already performs strongly
without consultation, leaving little headroom, and its remaining consultation-value
structure is difficult to learn from the small training set. Section~\ref{subsec:xfer}
shows that the failure is specific to learning Sol's value structure rather than
to allocation for this executor.

\paragraph{Takeaway.}
\method{} improves allocation by considering consultations jointly
and accounting for valuable later opportunities, while Local Value remains competitive in several settings.

\section{How Should Consultation Value Be Learned?}
\label{sec:alt_learning}

\method{} directly regresses total executor-relative consultation value without imposing an explicit factorization. 
We additionally evaluated explicit local-plus-interaction decompositions, variants that incorporate local-value estimates as auxiliary inputs or losses, relative and ranking-based objectives, and simpler predictors. 
Complete ablation results are presented in Appendix~\ref{appendix:add_results}.

\paragraph{Direct total-value regression is a strong default.}
A matched additive control under Sonnet~5 shows that, with set-level
supervision and preprocessing fixed, removing the additive constraint
significantly improves Animals at $B=3$ (+0.25 NG) and Meeting Planning
at $B=2$ (+0.12 NG), with no significant differences elsewhere
(Appendix~\ref{appendix:matched_additive_control}).
Under the main training protocol, alternatives significantly outperform
direct regression in 3 of 147 benchmark--budget comparisons and underperform it in 22 with Sonnet 5. And with GPT-6.1 Sol, the corresponding counts are 13 and 14.
Four of the five explicit decompositions remain significantly worse on Meeting Planning at $B=4$ under Sonnet 5.
A possible source of difficulty is error propagation: errors in estimated singleton values can propagate to predictions for every set containing the corresponding consultation. Direct regression learns the value of the complete allocation without requiring separate singleton estimates.

\paragraph{Additional structure helps in specific settings, while ensembling in general improves robustness.}
The benefits of local-value auxiliaries and relative or ranking-based objectives depend on the task and executor.
Under GPT-6.1~Sol, several alternatives outperform direct regression on Trip Planning at $B=1$, Flight Recommendation, and Meeting Planning.
Ensembling improves robustness and significantly helps several sequential settings. We therefore use ensemble \method{} as the default allocator throughout the main experiments.

\begin{figure*}[ht]
    \centering
    \begin{subfigure}[t]{0.32\textwidth}
        \centering
        \includegraphics[width=\linewidth]{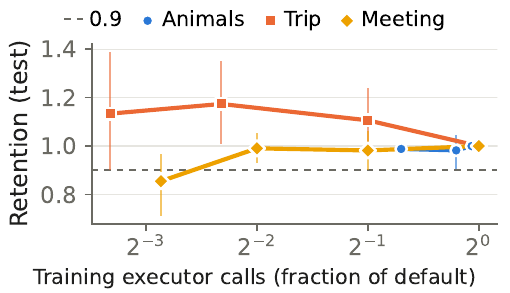}
        \caption{MC supervision}
        \label{fig:mc_efficiency}
    \end{subfigure}
    \hfill
    \begin{subfigure}[t]{0.32\textwidth}
        \centering
        \includegraphics[width=\linewidth]{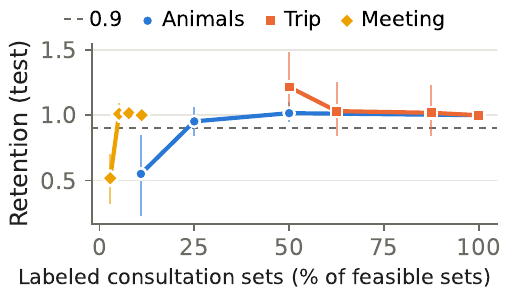}
        \caption{Subset supervision}
        \label{fig:subset_efficiency}
    \end{subfigure}
    \hfill
    \begin{subfigure}[t]{0.32\textwidth}
        \centering
        \includegraphics[width=\linewidth]{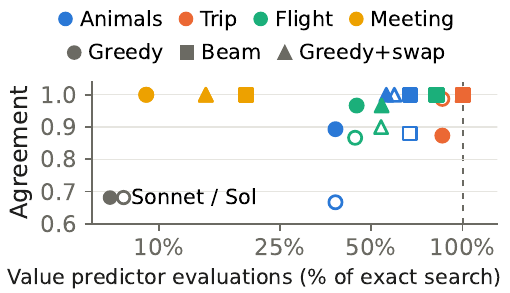}
        \caption{Approximate search}
        \label{fig:search}
    \end{subfigure}

    \caption{
    \textbf{Efficiency of \method{} at each benchmark's largest evaluated budget.}
(a) Retention relative to default supervision versus training executor calls,
reported as a fraction of default. $R_{\rm train}$ controls MC samples per
label. When hidden responses are marginalized, one sample may require multiple
calls.
(b) Retention as labeled consultation sets decrease. Points average over three
nested subset-sampling seeds. Error bars in (a)--(b) are paired 95\% CIs, and for (b), they use per-task means across seeds.
(c) Agreement with exact allocation versus value-predictor evaluations used by
approximate search. Full results are in Appendix~\ref{appendix:scalability}.
    }
    \vspace{-10pt}
    \label{fig:efficiency}
\end{figure*}

\section{How Practical and Reusable Is the Method?}
\label{sec:practicality}

In this section, we study the practical cost of \method{} and whether a value predictor trained for one executor can be reused with another. We first vary the amount of offline supervision and deployment time consultation set search, then evaluate cross-executor transfer.

\subsection{Efficiency}
\label{subsec:efficiency}

For the supervision ablations, we report retention, the reduced supervision allocator's gain ($\bar{U}$) over random ($\bar{U}_{\mathrm{random}}$) relative to that of the default supervision allocator ($\bar{U}_{\mathrm{default}}$):
\begin{align*}
    \mathrm{Retention}
=
(\bar U-\bar U_{\mathrm{random}})/
     (\bar U_{\mathrm{default}}-\bar U_{\mathrm{random}}).
\end{align*}
Since the default supervision predictor is itself learned rather than an upper bound, retention may exceed 1 when reduced supervision performs better on the test set. We report retention unclipped with paired 95\% confidence intervals. 
Values near one indicate similar gains to default supervision.

\paragraph{MC supervision.}
As shown in Figure~\ref{fig:mc_efficiency}, \method{} remains effective with fewer training executor
calls. At their largest budgets, Animals and Trip retain at least 90\% of the default supervision gain using 62\% and 10\% of the default training calls, respectively. Meeting retains at least 93\% across budgets using 25\% of the default calls, with no significant loss.

\paragraph{Subset supervision.}
As shown in Figure~\ref{fig:subset_efficiency}, \method{} also does not require labels for all feasible consultation sets.
Meeting Planning retains at least 93\% of the default supervision gain using
20 labeled sets per task instead of 43 by default
(5.2\% versus 11.1\% of the 386 feasible sets), while a quarter of the
Animals sets retains at least 95\%. Joint supervision nevertheless matters:
singleton-only training reduces retention to 0.52 on Meeting at $B=4$ and
0.55 on Animals at $B=3$, suggesting that broader set coverage is more useful
than highly precise supervision over only a few sets.

\paragraph{Approximate subset search.}
As shown in Figure~\ref{fig:search}, exact allocation grows combinatorially with the number of candidates, but approximate search is nearly lossless. Greedy selection followed by single-swap local search recovers the exactly enumerated allocation on $90$--$100\%$ of test tasks, and beam search is similarly accurate. On Meeting Planning at $B=4$, greedy+swap matches exhaustive search under both executors using only $55$ predictor evaluations instead of $386$. Even when approximate search selects a different set, the resulting normalized-gain loss is at most $0.06$ in our experiments.

\subsection{Cross-Executor Transfer}
\label{subsec:xfer}

As shown in Table~\ref{tab:xfer}, consultation value is executor-relative, but supervision can still transfer when executors induce similar allocation preferences.
Transfer is essentially lossless on Trip Planning and TB-D, but fails sharply on Animals, where the executor-specific allocators choose different consultation sets on $93\%$ of test episodes. 
Transfer can also be asymmetric. 
On Meeting Planning, Sonnet-trained \method{} improves Sol from $0.82$ to $0.91$, whereas Sol-to-Sonnet transfer reduces gain from $1.00$ to $0.89$. 
Flight Recommendation provides an informative exception. Although Sol-trained \method{} fails on Flight Recommendation, the Sonnet-trained allocator reaches NG $0.25$ when deployed with Sol, significantly above Sol's own allocator and close to Sol prompting ($0.29$). 
Thus, cross-executor supervision can be useful when value structure transfers, but does not eliminate the need for executor-specific learning.

\begin{table}[t]
\centering
\scriptsize
\setlength{\tabcolsep}{3pt}
\begin{tabular}{lccccc}
\toprule
& & \multicolumn{2}{c}{\textbf{Target: Sol}} 
& \multicolumn{2}{c}{\textbf{Target: Sonnet}} \\
\cmidrule(lr){3-4}\cmidrule(lr){5-6}
\textbf{Benchmark} & $P(\mathrm{diff})$
& Own & Sonnet$\rightarrow$Sol
& Own & Sol$\rightarrow$Sonnet \\
\midrule
Animals & 0.93 & 0.40 & 0.15$^{\ast}$ & 0.58 & 0.14$^{\ast}$ \\
Trip Planning   & 0.22 & 0.80 & 0.85              & 0.83 & 0.84 \\
Meeting & 0.50 & 0.82 & 0.91$^{\ast}$     & 1.00 & 0.89$^{\ast}$ \\
Flight Rec.  & 0.87 & 0.00 & 0.25$^{\ast}$     & 0.33 & 0.08 \\
TB-D    & 0.13 & 0.97 & 0.99              & 1.00 & 1.00 \\
\bottomrule
\end{tabular}
\caption{
Cross-executor transfer at the main-result budgets. 
\emph{Own}: \method{} trained on the target executor; arrows denote transfer from the other executor. 
$P(\mathrm{diff})$ is the fraction of test episodes where the two executor-specific allocators choose different sets.
$^{\ast}$: transferred versus own differs significantly (paired $95\%$ bootstrap).
Full transfer diagnostics are in Appendix~\ref{appendix:xfer}.
}
\vspace{-10pt}
\label{tab:xfer}
\end{table}

\section{Related Work}
\label{sec:related_work}

\paragraph{Human--AI deferral and scarce oversight.}
Learning-to-defer and algorithmic-triage methods decide when tasks should be handled by AI or transferred to human experts, including settings with finite human workload or oversight budgets \citep{madras2018predict,raghu2019algorithmic,mozannar2020consistent,mozannar2023whoshouldpredict,alves2024deccaf,zavattari2026onehuman,zhang2026calibration}. Related interactive imitation-learning and human-advice methods allocate limited human interventions during sequential execution \citep{hoque2022thriftydagger,zhang2022humans,min2025selfregulation}. In our setting, the human does not replace the agent: a fixed LLM remains the executor, while scarce human capacity supplies information, judgment, or authority. We therefore optimize the downstream \emph{executor-relative value} of consultation, including interactions among consultation opportunities.

\paragraph{Clarification and Value of Information.}
Prior work studies when and what LLM agents should ask users to clarify ambiguous intent, preferences, or missing information \citep{andukuri2024stargate,zhang2024clamber,zhang2025futureturns,huang2025proactive,patel2025adapt,wu2025collabllm,ross2025when2call,suri2026structured,gulati2026ask}. Most closely, \cite{dong2026voi} formulate human--agent communication through Value of Information (VoI), reasoning over latent state, possible responses, and downstream utility at inference time. Their released estimator uses one-step lookahead over a finite answer space. \method{} instead learns executor-relative consultation value from offline executor outcomes and allocates a shared human budget across batch or sequential decisions without response enumeration during allocation. Related learned metareasoning allocates limited internal computation by expected value \citep{callaway2018learning}. We instead allocate external human decision capacity.

\paragraph{Search and amortized computation.}
LLM agents increasingly use search and planning to improve reasoning and interaction \citep{hao2023rap,zhou2024lats,yu2025exact}. ExACT uses Reflective-MCTS to generate stronger behavior and distills search experience into a cheaper learned policy \citep{yu2025exact}. We similarly amortize expensive offline computation over \emph{human-consultation choices} rather than environment actions: executor rollouts provide value supervision offline, while deployment requires no rollout-based value estimation and can use inexpensive approximate subset search.

\section{Conclusion}
\label{sec:conclusion}

We study how agentic systems should allocate scarce human information, judgment, or authority across competing decisions. We formulate this as executor-relative human decision capacity allocation and introduce consultation value distillation, which learns from offline executor rollouts but requires no Monte Carlo rollouts at deployment. Across diverse tasks and two strong LLM executors, direct value learning provides a robust allocation strategy, suggesting a practical way to focus limited human attention where it has the greatest downstream impact.

\clearpage

\section*{Limitations}
\label{sec:limitations}

Our study relies on simulated or benchmark-defined human responses rather than live human participants, and the evaluated consultation budgets are modest. Broader validation is needed to establish how well the approach transfers to human populations and real organizational settings.
We also assume pre-specified consultation opportunities and use benchmark-specific pre-consultation features. Learning candidate consultations and transferable value predictors directly from raw task context remains open.

\section*{Use of AI Assistants}
Generative AI assistants, including ChatGPT and Claude Code,
were used for manuscript editing, code development, and
experimental execution. In particular, Claude Code implemented
the experimental pipelines and executed the experiments under
the authors' supervision. The authors formulated the research
questions and methodology, specified the experimental design,
and reviewed and validated the results. All AI-assisted work
was reviewed by the authors.

\bibliography{mybib}

\appendix

\section{Formal Problem Definition}
\label{appendix:formal_problem}

Let $x$ denote the task context and $\pi$ a fixed downstream executor.
A task unfolds over stages $t=1,\ldots,T$. At stage $t$, the allocator
observes history $h_t$, a set of currently available consultation
opportunities $C_t(h_t)$, and remaining human budget $b_t$. It chooses
$S_t$ from the feasible action family $\mathcal A_t(h_t,b_t)$, with
\[
S_t \subseteq C_t(h_t),
\qquad |S_t| \le b_t .
\]
The resulting human responses and environment evolution produce the next
history $h_{t+1}$, with remaining budget $b_{t+1}=b_t-|S_t|$.
We assume unit consultation cost throughout this work.

Let $U_\pi$ denote terminal task utility under executor $\pi$. Define
\begin{align*}
V^*_{\pi,T+1}(h_{T+1},b_{T+1})
    = \mathbb{E}[U_\pi \mid h_{T+1}],
\end{align*}
and, recursively, for $t=T,\ldots,1$,
\begin{align*}
   Q^*_{\pi,t}(S \mid h_t,b_t)
=
\mathbb{E}\!\left[
V^*_{\pi,t+1}(h_{t+1},b_t-|S|)
\mid h_t,S
\right],
\end{align*}
\begin{align*}
V^*_{\pi,t}(h_t,b_t)
=
\max_{S\in\mathcal A_t(h_t,b_t)}
Q^*_{\pi,t}(S\mid h_t,b_t).
\end{align*}
The expectation integrates over human responses, latent task state,
executor randomness, and environment transitions as applicable.

The continuation-aware value of consulting on $S$ relative to the empty
action at stage $t$ is
\[
F_{\pi,t}(S\mid h_t,b_t)
=
Q^*_{\pi,t}(S\mid h_t,b_t)
-
Q^*_{\pi,t}(\emptyset\mid h_t,b_t).
\]
Because $Q^*_{\pi,t}$ includes optimal allocation of the remaining budget,
$F_{\pi,t}$ captures both the immediate benefit of human information and
the opportunity cost of consuming human capacity now.

\paragraph{Batch setting.}
Batch allocation is the special case $T=1$, $h_1=x$,
$C_1=[n]$, and $b_1=B$. Let $Q_\pi^{\mathrm{batch}}(S\mid x)=\mathbb{E}[U_\pi\mid x,S]$ denote the expected terminal utility after revealing the consultations in $S$.
Since no future consultation decisions remain,
\begin{align*}
F^{\mathrm{batch}}_\pi(S \mid x)
    = Q^{\mathrm{batch}}_\pi(S \mid x)
- Q^{\mathrm{batch}}_\pi(\emptyset \mid x),
\end{align*}
and the optimal allocation is
\[
S^*
=
\arg\max_{|S|\le B}
F^{\mathrm{batch}}_\pi(S\mid x).
\]
No additivity is assumed, so this formulation allows complementarity and
redundancy among consultations.

\paragraph{Sequential setting.}
Feasible actions are singleton consultations (when $b_t>0$) and the empty
action. In TB-D and TB-H, $C_t(h_t)=\{q_t\}$: the empty action skips the
current decision while execution continues. In Animals, $C_t(h_t)$
contains the unasked questions; the empty action ends consultation, with
value given by the executor's expected utility from guessing immediately.
For a candidate $q$, the continuation-aware value is
\begin{align*}
&F^{\mathrm{seq}}_{\pi,t}(h_t,q,b_t)\\
&=
Q^*_{\pi,t}(\{q\}\mid h_t,b_t)
-
Q^*_{\pi,t}(\emptyset\mid h_t,b_t).
\end{align*}
With budget remaining, choose the candidate with maximal positive value,
or take the empty action if none is positive. Appendix~\ref{appendix:seq_targets}
gives the benchmark-specific recursions.

\section{Benchmark Adaptation and Construction}
\label{appendix:benchmark}

\paragraph{Common setup.} Each task contains candidate consultations that reveal components of human-private information unavailable to the executor. Predictors use only information available before consultation and never observe hidden answers or outcomes used for supervision or evaluation. In this section, we describe benchmark-specific task construction here.

\paragraph{Mixed 20Q Animals.} We adapt the Mixed 20Q animal game of \cite{dong2026voi}. Each episode contains 100 candidate animals and a uniformly sampled hidden animal. Six Yes/No/Maybe questions are generated without access to the hidden animal, after removing degenerate and highly redundant questions. Consulting reveals the answer to one question, and the executor must identify the hidden animal from the candidate pool and revealed answers. Utility is $2\Pr(\mathrm{correct})-1$. Animals supports both batch allocation over question sets and sequential allocation in which later questions may depend on earlier answers.

\paragraph{NaturalPlan Trip Planning.} We adapt Trip Planning tasks from NaturalPlan \cite{zheng2024naturalplan} in which the executor must construct a multi-city itinerary satisfying stay durations, direct-flight constraints, and dated commitments. We hide the dates of three commitments while retaining the event and city in the public task; each consultation reveals one hidden date. We retain tasks for which the hidden commitments materially constrain the feasible itineraries. Utility is one iff the produced itinerary satisfies all public and hidden travel constraints. Trip Planning is evaluated in the batch setting.

\paragraph{NaturalPlan Meeting Planning.} Each task asks the executor to schedule meetings with ten friends given travel times, availability windows, and minimum meeting durations. Every friend defines a candidate consultation. For eight friends, a private commitment shortens the publicly stated availability window while preserving individual feasibility. Consulting reveals the true window. Utility is the number of meetings successfully executed against the hidden true availabilities. Meeting Planning is evaluated in the batch setting.

\paragraph{Flight Recommendation.} We adapt the flight-recommendation setting of \cite{dong2026voi}, based on FLIGHTPREF \cite{lin2022inferringrewards}. Five observed flight choices constrain a latent user preference, and the executor must recommend one of three query flights. The five released preference questions form the consultation candidates. Consulting reveals the user's answer to the corresponding question. Utility is the released preference utility normalized between the expected worst and full-information choices. We evaluate Flight Recommendation in the batch setting.

\paragraph{ThinkingBox-D (TB-D).} We construct an IT-procurement workflow from ThinkingBox \citep{li2026thinkingbox}. The workflow contains decisions about supplier, billing, and delivery notification, with additional compliance decisions triggered when required by the supplier choice. Each decision has 3--6 listed options; the hidden correct option is sampled uniformly from the interior options, excluding the first and last, and the order is not shuffled. A colleague's note appears with probability $0.5$ and, when present, names the correct option with probability $0.6$. An incorrect note names an option uniformly sampled from those other than the correct and first-listed options. A consultation reveals the correct option for an available decision. Utility aggregates correctness over the workflow, with higher-weight decisions requiring additional follow-up actions. TB-D supports both batch and sequential allocation.

\paragraph{ThinkingBox-H (TB-H).} TB-H contains procurement, incident-response, and travel-booking workflows with nine or ten irreversible decisions. Earlier choices can create later obligations, enable options, make later answers inferable, or eliminate later decisions, creating continuation effects under a shared consultation budget. Each family has a coupled version in which these dependencies arise from earlier decisions and a matched control in which the same triggers depend on independent public randomness. One consultation opportunity is available per decision. TB-H is evaluated sequentially.

\begin{table}[t]
\centering
\small
\setlength{\tabcolsep}{3pt}
\begin{tabularx}{\columnwidth}{lX}
\toprule
Benchmark & Predictor information \\
\midrule
Animals &
Question partition statistics and pairwise redundancy. \\

Trip &
Stay/flight feasibility, hidden-date uncertainty, and features from a precomputed no-consultation draft. \\

Meeting &
Availability-window and travel-feasibility statistics. \\

Flight &
Query-flight variation, support-choice cues, and features from a precomputed no-consultation recommendation. \\

TB-D &
Decision type, weight, option count, workflow position, and public-note cues. \\

TB-H &
Workflow state, decision metadata, consultation history, and remaining budget. \\
\bottomrule
\end{tabularx}
\caption{Benchmark-specific predictor inputs. All features are available before consultation, and any LLM-derived features are precomputed offline. After predictor features are constructed, allocation requires no LLM or executor calls.}
\label{tab:predictor_features}
\end{table}

\paragraph{Predictor Features.} We summarize features of each benchmark used by the consultation value predictor in Table~\ref{tab:predictor_features}.

\subsection{Sequential Benchmark Coverage}
\label{appendix:seq_eligibility}

Adaptive evaluation from stored rollouts requires outcomes for the
consultation sets and response paths a policy can reach. Animals has this
coverage over its discrete answers; ThinkingBox uses the empirical workflow
replay described in Appendix~\ref{appendix:seq_targets}. Trip and Meeting
Planning sample hidden dates or availability windows independently for each
consultation set, leaving response paths without matching executor outcomes.
Evaluating adaptive allocation on these tasks would therefore require
additional rollouts.

Flight also supports adaptive evaluation, but development results showed
limited additional value over a fixed set (Table~\ref{tab:app_seq_headroom}).
Both references were selected on the label half and scored on the
independent evaluation half. The adaptive advantage is $0.009$--$0.017$ in
Flight's native normalized preference utility, versus $0.015$--$0.071$ in
Animals' native utility. Relative to the best fixed set's gain over random,
these gaps are $0.07$--$0.24$ and $0.33$--$1.30$, respectively. We therefore
retain Flight as a batch benchmark.

\begin{table}[t]
\centering
\scriptsize
\setlength{\tabcolsep}{3pt}
\begin{tabular}{llccc}
\toprule
Task & Executor & $B$ & Adaptive $-$ fixed & Ratio \\
\midrule
Animals & Sonnet & 2 & $0.017$ [$0.015$, $0.019$] & $0.59$ \\
 & & 3 & $0.057$ [$0.053$, $0.061$] & $1.30$ \\
 & Sol & 2 & $0.015$ [$0.013$, $0.017$] & $0.33$ \\
 & & 3 & $0.071$ [$0.068$, $0.075$] & $1.23$ \\
\midrule
Flight & Sonnet & 2 & $0.009$ [$-0.000$, $0.018$] & $0.07$ \\
 & & 3 & $0.012$ [$0.002$, $0.021$] & $0.13$ \\
 & Sol & 2 & $0.010$ [$0.002$, $0.018$] & $0.11$ \\
 & & 3 & $0.017$ [$0.011$, $0.023$] & $0.24$ \\
\bottomrule
\end{tabular}
\caption{Adaptive advantage over the best fixed set of at most $B$
consultations on non-test development tasks (Animals: 245; Flight: 113).
Gaps are in native utility with $95\%$ bootstrap intervals.
Ratio divides the gap by the best fixed set's gain over a random size-$B$
set. Animals uses $2\Pr(\mathrm{correct})-1$; Flight uses normalized
preference utility.}
\label{tab:app_seq_headroom}
\end{table}

\section{Prompts}
\label{appendix:prompts}

\paragraph{Our allocators.}
The value predictors and allocators use no prompts at deployment. They operate only on the precomputed features in Table~\ref{tab:predictor_features} and make no LLM or executor calls during allocation. Offline LLM calls are used to generate executor-rollout supervision and, where applicable, to precompute predictor features. For Animals, the released batch answer-simulation prompt is run with Sonnet~5 for both executor settings on every public question--candidate pair to estimate each question's Yes/No/Maybe partition of the 100-animal candidate pool. These simulated partitions are used to construct the question-partition features and select the candidate questions. They are distinct from, and do not reveal, the hidden animal's answers used during consultation.

\paragraph{Prompting baseline.} We report the exact prompt templates below. Text in braces is instantiated for each task. The prompting baseline uses the same model as the downstream executor, with medium reasoning effort, and we parse the final \texttt{ANSWER:} line as its allocation decision. 
For sequential benchmarks, the prompting baseline is queried independently at each decision point. 
We draw three samples per decision point, and the otherwise identical prompts are distinguished by a trailing identifier, \texttt{(sample 0)}, \texttt{(sample 1)}, or \texttt{(sample 2)}. The TB-H prompt exposes only information available to the agent and does not describe the benchmark's hidden-information mechanism.

\begin{promptbox}{Batch Animals}
You are about to play a game of 20 Questions. A hidden animal has been chosen uniformly at random from this list of 100 candidate animals:
{comma-separated candidate animals}

Before you must guess, you may ask exactly {B} of the following 6 questions. Each will be answered truthfully about the hidden animal with Yes, No, or Maybe. After seeing the answers you make ONE guess from the list above, and you score only if the guess is exactly right.

1. {question 1}
...
6. {question 6}

Which {B} question(s) should you ask to maximise your chance of guessing the animal correctly? Consider how well each question splits the candidates and how the questions combine.

Answer with exactly {B} question number(s) on the last line, formatted as: ANSWER: {1, ..., B}
\end{promptbox}

\begin{promptbox}{Trip Planning}
{native task description with the three hidden commitments' dates removed, the list of direct flights, and the native request "Find a trip plan of visiting the cities for {D} days by taking direct flights to commute between them."}

--- BEFORE YOU PLAN ---
The traveller has fixed dates for the following events, but the dates are NOT stated above. The stay in each city must cover its event's days; any dates consistent with a feasible itinerary are possible, and your plan counts as valid only if it is consistent with all of the true dates:
- {city 1}: {event 1}
- {city 2}: {event 2}
- {city 3}: {event 3}

You may ask the traveller for the exact dates of exactly {B} of these events before writing the plan; the other dates stay unknown. Which {B} event(s) should you ask about to maximise the chance that your plan is valid?

Answer with exactly {B} city name(s) on the last line, formatted as: ANSWER: {first B cities}
\end{promptbox}

\begin{promptbox}{Meeting Planning}
You are visiting San Francisco for the day and want to meet as many friends as possible.

Travel distances (in minutes):
{native travel-time table}

CONSTRAINTS: {native constraints: arrival location and time; each friend's location, stated window, and minimum meeting time}

--- BEFORE YOU PLAN ---
Exactly 8 of these 10 people have a private commitment you cannot see; the other 2 have none. For each of the 8, the real availability window is narrower than shown: a uniformly random 50\%-100\% of their visible slack (window length minus the required meeting time) is removed, from either the start (they cannot begin until later) or the end (they must leave earlier), each side equally likely. If your plan uses the shown window and the real one is narrower, that meeting fails when you arrive.

Before committing to a schedule you may consult exactly {B} of the 10 people; consulting a person reveals their TRUE availability window. Which {B} person/people should you consult to maximise the number of meetings you actually achieve? Think about how each person's availability interacts with your route and with the other people.

Answer with exactly {B} name(s) on the last line, formatted as: ANSWER: {first B names}
\end{promptbox}

\begin{promptbox}{Flight Recommendation}
{released executor prompt: instructions and output format, the five support rounds with the user's choices, and the options of the new round}

--- BEFORE YOU RECOMMEND ---
You will have to choose the best option for the NEW round for this user. Your choice is scored by how well it matches the user's true preferences (which generated the support-history choices above). Before choosing, you may ask the user exactly {B} of the following clarifying questions and see the answers:
1. Q: Do you prefer a flight with a lower price or a higher price? (A) Lower price (B) Higher price (C) No preference
2. Q: Do you prefer a flight with stops or a direct flight? (A) Direct flight (B) Flight with stops (C) No preference
3. Q: Do you prefer a flight with a short layover or a long layover? (A) Short layover (B) Long layover (C) No preference
4. Q: Do you prefer to arrive a little before your meeting or a lot before your meeting? (A) A little before (B) A lot before (C) No preference
5. Q: Which airline do you prefer? (A) American, (B) Delta, (C) JetBlue, (D) Southwest (E) No preference

Which {B} question(s) should you ask to make the best recommendation?

Answer with exactly {B} question number(s) on the last line, formatted as: ANSWER: {1, ..., B}
\end{promptbox}

\paragraph{TB-D: shared text.} Both TB-D prompts describe each decision with the block the executor itself sees (example below), and both include the description of the generator, \texttt{\{MECHANISM\}}, shown after it.

\begin{promptbox}{TB-D decision block (example)}
decision_type: order_vendor
ticket: TKT-01 -- Order the replacement docking stations
detail: Place the order (one lot) with the supplier procurement has standardised on for this part. 3 suppliers can supply it (`list_vendors`); the ticket does not record which one procurement uses. Supplier policy: an order placed with a supplier that is not yet onboarded automatically opens two compliance sub-tickets -- a security review and a contract signatory (TKT-02 and TKT-03) -- which must then be completed. A colleague added a note: "I think it was VEN-12 last time -- not certain."
visible cue points to: VEN-12
candidate options:
  1. VEN-11 (name Supplier A, quote \$800)
  2. VEN-12 (name Supplier B, quote \$800)
  3. VEN-13 (name Supplier C, quote \$800)

[VISIBLE STATE]
decision weight: 1
position in workflow: 1
is a compliance obligation: False
\end{promptbox}

\begin{promptbox}{TB-D \{MECHANISM\}}
How the hidden answers were generated: every decision has exactly one correct option. A colleague note ("visible cue") is present for about half of the decisions; when present it names the correct option with probability 0.6, and otherwise names a wrong option, chosen uniformly among the options other than the correct one and the first-listed one. Without a note, the correct option is equally likely to be any listed option except the first-listed and the last-listed ones (the first- and last-listed options are never correct). For the supplier order, the correct supplier is already onboarded with probability 0.8 and each other supplier with probability 0.5 (onboarding status is hidden here); ordering from a supplier that is not onboarded opens the two compliance sub-tickets (security review, contract signatory), which then also have to be completed and scored. A correct decision earns its weight in points; compliance obligations that are triggered and answered wrongly cost their weight.
\end{promptbox}

\begin{promptbox}{Batch TB-D ($B=1$)}
[WORLD / TASK CONTEXT]
IT procurement: several sub-tickets, each under-specified. You will work through the five decisions below in order (d1 first).

=== DECISION d1 ===
{decision block d1}
...
=== DECISION d5 ===
{decision block d5}

--- BEFORE YOU START ---
{MECHANISM}

You may consult a human about exactly 1 of these decisions before you start; the human tells you the correct option for it. Which decision should you consult on to maximise your total points?

Answer with exactly 1 decision id on the last line, formatted as: ANSWER: d1
\end{promptbox}

\begin{promptbox}{Sequential Animals (asked at every node with budget left)}
You are playing a game of 20 Questions. A hidden animal has been chosen uniformly at random from this list of 100 candidate animals:
{comma-separated candidate animals}

You may ask up to {B} of the following 6 questions, ONE AT A TIME, and you see each answer (Yes, No, or Maybe, answered truthfully about the hidden animal) before choosing the next. After your questions you make ONE guess from the list above, and you score only if the guess is exactly right.

1. {question 1}
...
6. {question 6}

Questions asked so far and their answers:
- Q{i} "{question i}" -> {answer}      (or: - none yet)

You have {b} question(s) left. Which question should you ask NEXT to maximise your chance of guessing the animal correctly, given the answers so far? You may also STOP and guess now.

Answer on the last line with exactly one of: ANSWER: {i} (one per unasked question), ANSWER: STOP
\end{promptbox}

\begin{promptbox}{Sequential TB-D (asked at every arriving decision with budget left)}
[WORLD / TASK CONTEXT]
IT procurement: several sub-tickets, each under-specified. Decisions arrive one at a time in a fixed workflow order: d1 (supplier order), then d2 and d3 (compliance sub-tickets, which open only if the ordered supplier is not onboarded), then d4, then d5. You may consult a human at most {B} time(s) in this whole request; you have {b} consultation(s) left. When you consult, the human tells you the correct option for that decision.

{MECHANISM}

Compliance status: {one of: "not yet known (it is determined when the supplier order is placed)" | "the order went to a supplier that was NOT onboarded, so the two compliance sub-tickets (d2, d3) are open" | "the order went to an onboarded supplier, so no compliance sub-tickets were opened"}.

Decisions handled so far:
=== {d} ({CONSULTED | decided yourself}) ===
{decision block}      (one per handled decision, or: - none)

=== CURRENT DECISION {d} ===
{decision block}

=== UPCOMING DECISION {d} ===
{decision block}      (one per visible upcoming decision, or: (none currently visible))

Should you consult the human on decision {d} NOW (using one of your remaining consultations), or decide it yourself and save the consultation for a later decision? Answer on the last line with exactly: ANSWER: CONSULT  or  ANSWER: SKIP
\end{promptbox}

\begin{promptbox}{TB-H (asked at every arriving ticket with budget left)}
You are an AI agent completing a multi-ticket {family} request. Tickets arrive one at a time in a fixed workflow order (this is ticket {i} of at most {N}); some later tickets open only depending on earlier decisions. For each ticket you either decide yourself or consult the responsible human, who then tells you the correct option. You may consult at most {B} time(s) in this whole request; you have {b} consultation(s) left. A wrong decision loses that ticket's points and can open extra follow-up tickets.

Tickets handled so far:
- {ticket id} {summary}: {CONSULTED | decided yourself}      (or: - none)

Current visible state: {visible workflow state, as JSON}

CURRENT TICKET:
{ticket id} -- {summary}
{detail}
Options:
  - {option}: {label}
Possible consequences: {visible consequences of the options}
Question you could ask {responsible human}: {question}

Should you consult the human on THIS ticket now (using one of your remaining consultations), or decide it yourself and save the consultation for a later ticket? Answer on the last line with exactly: ANSWER: CONSULT  or  ANSWER: SKIP
\end{promptbox}

\section{Benchmark Statistics}
\label{appendix:benchmark_stats}

Table~\ref{tab:app_label_stats} summarizes the consultation-value
supervision for each benchmark and set size, including the number of feasible and labeled sets per task, executor samples per label, and label reliability.
Training and evaluation labels are estimated from independent executor
samples, and hidden task state is marginalized exactly where possible.
The deployed \method{} predictor requires no executor rollouts.

Here, \emph{Reliability} $\rho$ measures agreement between the two independent halves of the value supervision. For the batch setting benchmarks, it is the Pearson correlation between independent estimates of $F_\pi^{\mathrm{batch}}(S\mid x)$, pooled over task--set pairs $(x, S)$ with fixed $|S|$. For the sequential setting benchmarks, the table reports reliability of the underlying fixed-set value estimates from which the continuation-aware targets $F_{\pi,t}^{\mathrm{seq}}(\{q_t\} \mid h_t,b_t)$ are constructed. The sequential targets themselves are computed independently from the two rollout halves. For $|S|=0$, we instead correlate the two estimates of $Q_\pi^{\mathrm{batch}}(\emptyset\mid x)$ (or the corresponding no-consultation outcome estimate).

\begin{table*}[t]
\centering
\small
\setlength{\tabcolsep}{5pt}
\begin{tabular}{lccccccccc}
\toprule
 & & \multicolumn{2}{c}{Sets per task} & \multicolumn{2}{c}{Samples per label} & \multicolumn{2}{c}{$\rho_{\rm train}$} & \multicolumn{2}{c}{$\rho_{\rm test}$} \\
\cmidrule(lr){3-4}\cmidrule(lr){5-6}\cmidrule(lr){7-8}\cmidrule(lr){9-10}
Benchmark & $|S|$ & Feasible & Labeled & $R_{\rm train}$ & $R_{\rm test}$ & Sonnet~5 & GPT-6.1~Sol & Sonnet~5 & GPT-6.1~Sol \\
\midrule
Animals & 0 & 1 & 1 & 8 & 8 & 0.88 & 0.99 & 0.84 & 1.00 \\
 & 1 & 6 & 6 & 8 & 8 & 0.91 & 0.97 & 0.93 & 0.97 \\
 & 2 & 15 & 15 & 3 & 3 & 0.87 & 0.94 & 0.85 & 0.94 \\
 & 3 & 20 & 20 & 2 & 2 & 0.85 & 0.92 & 0.81 & 0.92 \\
 & 4 & 15 & 15 & 1 & 1 & 0.76 & 0.88 & 0.70 & 0.89 \\
 & 5 & 6 & 6 & 1 & 1 & 0.76 & 0.89 & 0.71 & 0.89 \\
 & 6 & 1 & 1 & 1 & 1 & 0.76 & 0.89 & 0.63 & 0.90 \\
\addlinespace
Trip & 0 & 1 & 1 & 10 & 10 & 0.97 & 0.99 & 0.98 & 0.98 \\
 & 1 & 3 & 3 & 10 & 10 & 0.93 & 0.96 & 0.93 & 0.97 \\
 & 2 & 3 & 3 & 10 & 10 & 0.90 & 0.94 & 0.91 & 0.94 \\
 & 3 & 1 & 1 & 10 & 10 & 0.84 & 0.96 & 0.80 & 0.96 \\
\addlinespace
Meeting & 0 & 1 & 1 & 32 & 50 & 0.98 & 1.00 & 0.99 & 0.99 \\
 & 1 & 10 & 10 & 12 & 12--50 & 0.65 & 0.89 & 0.71 & 0.90 \\
 & 2 & 45 & 9 & 8 & 12--50 & 0.72 & 0.90 & 0.81 & 0.92 \\
 & 3 & 120 & 9 & 8 & 12--50 & 0.77 & 0.91 & 0.82 & 0.92 \\
 & 4 & 210 & 14 & 8 & 12--50 & 0.76 & 0.91 & 0.84 & 0.93 \\
\addlinespace
Flight & 0 & 1 & 1 & 4 & 4 & 0.98 & 0.96 & 0.92 & 0.90 \\
 & 1 & 5 & 5 & 4 & 4 & 0.91 & 0.86 & 0.84 & 0.69 \\
 & 2 & 10 & 10 & 2 & 2 & 0.92 & 0.89 & 0.85 & 0.78 \\
 & 3 & 10 & 10 & 2 & 2 & 0.94 & 0.91 & 0.87 & 0.81 \\
\addlinespace
TB-D & 0 & 1 & 1 & 8 & 8 & 0.89 & 0.90 & 0.80 & 0.81 \\
 & 1 & 5 & 5 & 8 & 8 & 0.86 & 0.89 & 0.84 & 0.89 \\
 & 2 & 10 & 10 & 8 & 8 & 0.88 & 0.90 & 0.75 & 0.84 \\
\addlinespace
TB-H & 0 & 1 & 1 & 4 & 4 & 0.89 & 0.89 & 0.88 & 0.84 \\
 & 1 & 9--10 & 9--10 & 4 & 4 & 0.84 & 0.90 & 0.80 & 0.85 \\
 & 2 & 36--45 & 36--45 & 4 & 4 & 0.89 & 0.91 & 0.87 & 0.88 \\
\bottomrule
\end{tabular}
\caption{\textbf{Consultation-value supervision and label reliability} by benchmark and
set size.
\emph{Labeled} is the number of sets labeled per training task; all
feasible sets are labeled on test tasks.
$R_{\rm train}$ and $R_{\rm test}$ denote independent Monte Carlo samples used to estimate training and evaluation labels; a sample may require multiple executor calls when marginalizing over hidden responses.
Animals enumerates all induced answer patterns; Flight enumerates answer combinations with probability at least $10^{-4}$; Trip, Meeting, and ThinkingBox additionally average over sampled hidden states or completions.
Sequential Animals reuses the batch labels with $|S|\le3$; TB-D
$|S|=2$ is used only in the sequential setting.}
\label{tab:app_label_stats}
\end{table*}

\paragraph{Monte Carlo samples and executor calls.}
$R_{\rm train}$ counts independent Monte Carlo samples rather than raw
executor calls. For Trip and Meeting, each sample requires one executor call. For Animals, one sample enumerates the answer patterns induced by the consultation set, requiring on average $1$, $2.8$, $7.0$, and $14.1$ calls for $|S|=0,1,2,3$, respectively (and $23$--$42$ for larger sets). For Flight, one sample evaluates each answer combination with probability at least $10^{-4}$, requiring on average $1$, $3.3$, $9.6$, and $24.1$ calls for $|S|=0,1,2,3$. Thus $R_{\rm train}=1$ should not be interpreted as one executor call per label on these benchmarks.

\paragraph{Output scoring and failures.}
Executor outputs are never discarded because they are malformed or
unsuccessful. They are scored by the benchmark-specific evaluator.
Parsing failures are negligible on most benchmarks. The main exception is Trip Planning, where unparseable or incomplete itineraries increase with the number of revealed commitments and reach $19.8\%$ for Sonnet~5 and $25.9\%$ for GPT-6.1~Sol at $|S|=3$. These outputs are scored as invalid itineraries. In TB-H, $1.4\%$ of Sonnet~5 and $8.9\%$ of GPT-6.1~Sol episodes end before completing the workflow and are scored using the outcomes actually achieved. Failures attributable solely to infrastructure rather than the executor are rerun.

\section{Evaluation and Implementation Details}
\label{appendix:metrics_eval}

\subsection{Normalization References and Evaluation Setup}
\label{appendix:ref_eval}

As mentioned in Section~\ref{subsec:metrics}, we report normalized gain (NG) using references
$\bar{U}_{\mathrm{random}}$ and $\bar{U}_{\mathrm{oracle}}$. We specify these references here.
Unless noted otherwise, oracle selection and final scoring use the two
independent label halves described in Appendix~\ref{appendix:benchmark_stats}.

Let $\widehat{U}_{\pi,L}(S\mid x)$ and $\widehat{U}_{\pi,E}(S\mid x)$ denote the two independent label-half and evaluation-half estimates of the expected terminal utility obtained when consultation set $S$ is provided to executor $\pi$.

\paragraph{Batch benchmarks.}
For Animals, Trip Planning, Meeting Planning, Flight Recommendation, and
batch TB-D, $\bar{U}_{\mathrm{random}}$ is the exact expected utility of a uniformly
random size-$B$ consultation set, averaged over all feasible sets of that
size. $\bar{U}_{\mathrm{oracle}}$ is an \emph{ex-ante executor-relative subset oracle}:
for each test task, it selects
\begin{align*}
    S^\star
=
\arg\max_{|S|\le B}
\widehat U_{\pi,L}(S\mid x)
\end{align*}
using the label half and is scored using the independent evaluation estimate
$\widehat{U}_{\pi,E}(S^\star\mid x)$.
Ties in the label-half utility are averaged.
Because the oracle is selected from noisy utility estimates, it is a
normalization reference rather than a strict upper bound.

\paragraph{Sequential Animals.}
$\bar{U}_{\mathrm{random}}$ is the same uniformly random size-$B$ question set used in
the batch setting, committed to before any answers are observed.
$\bar{U}_{\mathrm{oracle}}$ is the ex-ante optimal adaptive allocator: at each history,
it selects the next consultation (or stops) according to the continuation
values $Q^*_{\pi,t}$ defined in Section~\ref{sec:problem}.
The allocator is constructed using the label half and scored on the
independent evaluation half.

\paragraph{Sequential TB-D and TB-H.}
Because consultation opportunities depend on the evolving workflow, a
uniformly random size-$B$ set is not well defined. We therefore use
\emph{no consultation} as $\bar{U}_{\mathrm{random}}$. $\bar{U}_{\mathrm{oracle}}$ is the optimal
adaptive allocator computed by dynamic programming over the evaluation
completions, using the same observable workflow state and budget as the
learned allocator. This is an empirical reference for those evaluation
completions; for TB-H it is defined within the approximate replay process
described in Appendix~\ref{appendix:seq_targets}.

\paragraph{Evaluation independence.}
Learned predictors are trained only on non-test tasks and never use the
evaluation-half outcomes. Batch methods are scored on the independent
evaluation half. Sequential Animals uses the corresponding independent
rollout half, while TB-D and TB-H are scored on held-out executor
completions. Model choices and hyperparameters are frozen before final test
evaluation.

\subsection{Sequential Target Construction}
\label{appendix:seq_targets}

We compute sequential targets by dynamic programming over stored executor
outcomes. Animals chooses any unasked question or stops; TB-D and TB-H
consult or skip the current workflow decision. The recursions below
specify these distinct action spaces and require no new executor calls.

\paragraph{Observable histories and outcome halves.}
A node contains the observable history $h$ and remaining budget $b$.
Its consistent set $K(h)$ contains the hidden states compatible with the
observations under the actions in $h$. States within a half have equal
prior weight. Animals enumerates all 100 candidate animals and splits
executor samples into independent halves. TB-D samples 16 hidden
completions (correct options and supplier onboarding), with eight per half;
TB-H samples eight completions, with four per half. A completion is hidden
from the allocator. The stored terminal utility $u(C,k)$ for reveal set
$C$ and completion $k$ is posterior-averaged on TB-D and the observed
checkpoint sum on TB-H. Label construction and evaluation use separate
halves as specified in Appendix~\ref{appendix:ref_eval}.

The DP selects one action for each observable history, averaging over its
consistent states before comparing actions. It never selects an action for
an individually identified hidden state. In TB-D, the history records
workflow position, past consult/skip choices, and the compliance trigger
once observed. Answer contents and executor choices are not conditioning
variables: correct options are independent across decisions, and the
trigger captures the dependency used by this allocator. In TB-H, the
history additionally includes consultation answers, submitted executor
choices, and visible workflow state; an answer may indicate that none of
the offered options is correct.

\paragraph{ThinkingBox recursion.}
Let $i$ be the next workflow position, $C$ the consulted set, and $K$ the
consistent completions. Before choosing an action, partition $K$ into
groups $K'$ with the same observable arrival status (and, for TB-D, trigger
status). With $b=B-|C|$,
\[
V(i,C,K)=\sum_{K'}\frac{|K'|}{|K|}\,W(i,C,K').
\]
For an arriving decision,
$W(i,C,K')=\max_{a\in\mathcal A(b)}Q(i,C,K',a)$; otherwise
$W(i,C,K')=V(i+1,C,K')$.
Here $\mathcal A(b)=\{\mathrm{skip},\mathrm{consult}\}$ for $b>0$ and
$\{\mathrm{skip}\}$ otherwise. For action $a$, set $C_a=C\cup\{i\}$ if
consulting and $C_a=C$ if skipping. Partition the current consistent set
by the post-action observation $o$:
\begin{align*}
Q(i,C,K,a)
&=\sum_o\frac{|K_o|}{|K|}V(i+1,C_a,K_o),\\
V(n+1,C,K)
&=\frac{1}{|K|}\sum_{k\in K}u(C,k).
\end{align*}
For TB-D the post-action observation is empty; the trigger is observed at
the arrival step. At each arriving decision with budget remaining, let
$U_0=|K|^{-1}\sum_{k\in K}u(C,k)$ be utility with no further consultations.
The two supervised targets are
\begin{align*}
y^{\rm c}&=Q(i,C,K,\mathrm{consult})-U_0,\\
y^{\rm s}&=Q(i,C,K,\mathrm{skip})-U_0.
\end{align*}
Their difference is the estimated continuation-aware consultation value.
The myopic target instead consults at $i$ and uses no later consultations.

\paragraph{Reading stored workflows.}
Each episode $(k,C)$ runs the full workflow, revealing answers in $C$ only
when their decisions arrive. At position $i$, replay reads arrival from
$(k,C)$ and the post-action observation from $(k,C_a)$; subsequent replay
uses $(k,C_a)$ and terminal utility is $u(C_{\rm final},k)$.
For TB-D, the trigger is read from the no-consultation episode unless the
supplier decision was consulted, in which case its consulted episode
determines the trigger. Untriggered compliance decisions are removed from
the reveal set when reading utility. Across episodes without supplier
consultation, trigger rates agree exactly for Sol and have per-world
standard deviation $0.003$ for Sonnet ($0.11$ under binomial noise alone).

\paragraph{TB-H replay approximation.}
TB-H episodes for different reveal sets are independent executor runs that
share the hidden completion but not the execution prefix. After switching
episodes, earlier choices and visible workflow state can differ from those
already observed, and terminal utility is the final episode's full
checkpoint sum. TB-H training targets, adaptive-policy results, and the
adaptive normalization reference therefore use an approximate replay process;
exact equivalence to chronological trajectories is not assumed.
Prefix inconsistencies affect $45\%$ of \method{}'s Sonnet~5 test paths in
the uncoupled variant ($81/180$ across three training runs). Restricting to
paths without inconsistent switches for either \method{} or Local Value
yields a Sonnet~5 utility gap of $+0.53$ (95\% CI $[0.18,0.91]$), compared
with $+0.70$ over all paths. This post-hoc restriction changes the evaluated
subset and does not validate chronological execution. Missing arrivals
occur in $337$ of $472{,}875$ Sonnet and $363$ of $482{,}432$ Sol DP lookups;
these counts measure absent slots, not prefix consistency.

\paragraph{Chronological TB-H check.}
We directly assess the effect of replay using the frozen first training
run of \method{} and Local Value under both executors at $B=2$.
Each of the original 30 test worlds is evaluated on four evaluation-half
completions, with two fresh executor runs per policy and completion
(480 episodes per executor). At each arriving decision, the policy
receives the live workflow state and any requested answer is revealed
on arrival. Executor settings and utility scoring match the stored
workflows. All episodes pass checks of the observed states, consultation
decisions, revealed answers, and budget against the executor's trajectory.

With Sonnet~5, the chronological \method{}--Local Value utility gap is
$+0.86$ ($95\%$ CI $[0.40,1.37]$), versus $+0.78$
($[0.34,1.28]$) under replay. With GPT-6.1~Sol, the gaps are
$+0.71$ ($[0.38,1.02]$) and $+0.71$ ($[0.36,1.04]$), respectively.
The chronological-minus-replay changes in these gaps are $+0.09$
($[-0.04,0.23]$) and $+0.00$ ($[-0.09,0.11]$).
Intervals use 4{,}000 paired bootstrap resamples of test worlds after
averaging over completions and executor runs; differences are computed
before rounding. Sonnet~5 Local Value has lower chronological utility
than under replay ($-0.16$, $[-0.27,-0.06]$), so the check does not
establish equality of individual policy utilities.
It supports the \method{}--Local Value comparison for the tested policies.
The check covers only the first training run and the original test worlds;
Prompt and the adaptive normalization reference were not executed
chronologically. Training targets and the normalization reference
continue to use approximate replay.

\paragraph{Animals recursion and normalization.}
Let $S$ be the questions already asked, $a_S$ their observed answers, and
$Z(h)$ the animals consistent with those answers in the answer table.
Then $P(h)=|Z(h)|/N$, where $N=100$. The cached batch cells give the
probability-weighted stopping contribution
\[
c(S,Z)=\frac1N\sum_{z\in Z}
\left(2\widehat{\Pr}[g(S,a_S)=z]-1\right),
\]
where $g$ is the executor's final guess and $\widehat{\Pr}$ averages over
the chosen rollout half. Contributions add across answer branches:
\begin{align*}
\bar Q(S,Z,q)&=\sum_a\bar V(S\cup\{q\},Z_{q,a}),\\
\bar V(S,Z)&=\max\left\{c(S,Z),\max_{q\notin S}\bar Q(S,Z,q)\right\},
\end{align*}
where $Z_{q,a}=\{z\in Z:a_q(z)=a\}$ and only nonempty branches are
included. At $|S|=B$, $\bar V=c$. The target for candidate $q$ is
\[
y^{\rm A}(h,q)=\frac{\bar Q(S,Z,q)-c(S,Z)}{P(h)}.
\]
The numerator is a probability-weighted contribution, so division by
$P(h)$ produces the \emph{conditional} utility gain over stopping; the
conditional $Q$ values defined in Section~\ref{sec:problem} require no
further division. At deployment, the allocator selects the remaining
question with the highest positive ensemble score, or stops if none is
positive. The myopic target replaces $\bar V$ by $c$ in the branch sum.
Reachable training examples have equal weight within the groups $b=1$
and $b\ge2$, and the groups receive equal total loss weight.

\paragraph{Label-generation procedure.}
For each training task and outcome half:
\begin{enumerate}[leftmargin=*,itemsep=1pt,topsep=2pt]
\item Start at the empty history, full budget, and all states in the half.
\item In ThinkingBox, partition by observable arrival/trigger status and
update the history; advance past decisions that do not arrive.
\item At each reachable decision with budget left, compute the memoized
continuation values above and emit the observable features with their
consult/skip targets (ThinkingBox) or candidate-question targets (Animals).
\item Enumerate feasible actions, partition consistent states by each
possible observation, update history and budget, and recurse. Terminal
nodes use the stored utility; stopping in Animals uses its cached guess.
\end{enumerate}
Predictor inputs depend only on public task information and the observable
history. Stored outcomes and consistent hidden states are used to compute
targets, not as predictor inputs.

\paragraph{Worked example.}
Consider four equally likely animals and two binary questions whose
partitions cross, so both answers identify the animal. Suppose the
executor guesses one consistent animal correctly with zero or one answer,
and is always correct with both. Root stopping utility is $-0.5$.
After either answer to the first question, $P(h)=1/2$, stopping contributes
$c=0$, and asking the second contributes $\bar Q=1/2$. Thus its conditional
target is $(1/2-0)/(1/2)=1$, and the first question's continuation-aware
target is $1-(-0.5)=1.5$. Each decision conditions only on the observed
answer, not the identity of the hidden animal.

\subsection{Model Specifications and Training Configurations}
\label{subsec:model_training}

All learned consultation value predictors are lightweight MLP-based models trained from scratch. Architecture and training hyperparameters are fixed across executors, although supervision and reachable training states may differ. Training hyperparameters were selected using development data before test evaluation and are shared across benchmarks within each setting. Tables~\ref{tab:app_hparams} and \ref{tab:app_params} summarize the training configurations, input dimensions, and parameter counts.

\paragraph{Batch models.}
The value predictor $D_{\theta}$ uses a DeepSets architecture~\citep{zaheer2017deepsets}. Each consultation feature $z_i\in\mathbb{R}^{d_z}$ is encoded by a two-layer MLP ($d_z\!\to\!64\!\to\!64$) with ReLU activations. Embeddings are summed over the selected set $S$ and concatenated with task context $c_x\in\mathbb{R}^{d_c}$ and normalized set size $|S|/n$. A prediction head ($64+d_c+1\!\to\!64\!\to\!1$) estimates the set value. The local baseline independently scores each candidate using an MLP ($d_z+d_c\!\to\!64\!\to\!1$) with ReLU and dropout. Both models minimize squared error against the corresponding set or singleton consultation values, with targets scaled by their training standard deviation. \method{} also includes the empty set (target zero) in training.

\paragraph{Sequential models.}
The continuation-aware predictor and the myopic Local Value baseline use
two-hidden-layer MLPs ($d\!\rightarrow\!64\!\rightarrow\!64$) with ReLU
activations and dropout. Targets are constructed as specified in
Appendix~\ref{appendix:seq_targets}.

\paragraph{TB-D and TB-H.}
\method{} uses a shared encoder with two output heads,
$G^{\rm c}_\phi(h_t,q_t,b_t)$ and
$G^{\rm s}_\phi(h_t,q_t,b_t)$, supervised on $y^{\rm c}$ and $y^{\rm s}$:
\begin{align*}
\mathcal{L}^{\rm seq}(\phi)
=
\mathbb{E}\!\left[
    \bigl(G^{\rm c}_\phi-y^{\rm c}\bigr)^2
    +
    \bigl(G^{\rm s}_\phi-y^{\rm s}\bigr)^2
\right].
\end{align*}
Both targets are scaled by the training-set standard deviation of
$y^{\rm c}$. The common stopping reference cancels in
$G_\phi=G^{\rm c}_\phi-G^{\rm s}_\phi$. At deployment, we average this
difference across the five-model ensemble and consult iff the score is
positive and budget remains.

\paragraph{Sequential Animals.}
One output head predicts the conditional gain $y^{\rm A}(h,q)$ from
Appendix~\ref{appendix:seq_targets}, using squared error and target scaling
by its training-set standard deviation. Ensemble scores rank the remaining
questions against stopping.

\paragraph{Sequential training details.}
Inputs are standardized using training-set statistics. ThinkingBox budget
levels receive equal total loss weight; Animals uses the two budget groups
specified in Appendix~\ref{appendix:seq_targets}.

\paragraph{Local Value baseline.}
The sequential Local Value baseline predicts the gain from consulting now and then stopping consultation. Its five-model ensemble averages predictions in native utility units. With budget remaining, Animals asks the highest-scoring unasked question if its gain exceeds $\tau_B$, otherwise stopping; TB-D/TB-H consult the arriving decision iff its gain exceeds $\tau_b$, otherwise skipping. Thresholds are selected separately for each training run to maximize summed label-half policy utility on non-test tasks, including development tasks, then frozen for testing. Animals uses one threshold per initial budget $B$; ThinkingBox jointly selects thresholds for the remaining budgets $b$. The grids are $\{-\infty,0,0.005,0.01,0.02,0.04,0.08,\infty\}$ for Animals and $\{-\infty,0,0.5,1,1.5,2,2.5,3,4,\infty\}$ for ThinkingBox, with ties resolved in increasing order, lexicographically for threshold pairs.

\paragraph{Training and inference.}
All models are optimized with AdamW using the
configurations in Table~\ref{tab:app_hparams}.
Each allocator ensembles five independently initialized
models by averaging their predictions. We repeat
training three times with independent seeds and report
the mean and standard deviation across runs.

\paragraph{Executors and baselines.} On Animals, Trip, Meeting Planning, and Flight, executor rollouts use Sonnet~5 with extended thinking disabled and GPT-6.1~Sol at its minimum reasoning effort (low). On TB-D and TB-H, Sonnet~5 uses adaptive thinking at high effort, while GPT-6.1~Sol uses high effort for TB-D and medium effort for TB-H, with up to 32{,}000 output tokens. Label and evaluation halves of each benchmark use the same configuration. The prompting baseline uses the executor model with reasoning enabled (Sonnet~5: adaptive thinking at medium effort, up to 16{,}000 output tokens; GPT-6.1~Sol: medium reasoning effort, 8{,}000 visible output tokens) and three independent samples.

\paragraph{VoI adaptation.} We use the released value estimator of \cite{dong2026voi} (commit \texttt{27466a7}) with its one-step lookahead, changing only the LLM transport. Its belief model uses the executor model and settings for Animals and Flight. We adapt allocation to the benchmark's fixed candidate pool and budget: batch Animals and Flight rank candidates once at the empty history and select the top $B$; sequential Animals recomputes VoI for every remaining question after each observed answer and asks the highest-scoring question. Both settings spend exactly $B$ consultations, with no stopping rule and zero communication cost. Exact ties are averaged, and results are means over three independent runs.

Confidence intervals are $95\%$ percentile bootstrap intervals over test episodes with 4{,}000 resamples, resampling the method and both references jointly.

\paragraph{VoI with generated questions.}
We additionally evaluate the released Animals policy that proposes up to
three yes/no questions at each history and selects the one with highest
one-step VoI. This extends beyond the six predefined questions available
to the allocators in our main evaluation. We use the same 75 test episodes,
executor settings, and three independent runs. The common LLM wrapper
does not forward the released code's temperature arguments.
New questions are answered by the benchmark's oracle using a three-sample
majority, and outcomes requiring new executor calls are scored with fresh
rollouts under the original evaluation protocol. Utility is averaged over
all hidden animals and then over runs within each episode.
We evaluate the first $B=2,3$ questions and the root decision at $B=1$;
the released stopping rule with zero communication cost does not stop
before these budgets.

The following results use unclipped NG and the original normalization
references: batch references at $B=1$ and sequential references at
$B=2,3$. These references use predefined questions, so they do not bound
the performance of a policy that generates new questions. Brackets give
$95\%$ bootstrap confidence intervals for the question-generating policy.

\begin{center}
\footnotesize
\setlength{\tabcolsep}{3pt}
\begin{tabular}{@{}crr@{}}
\toprule
$B$ & \makecell{Predefined\\questions} & \makecell{Generated\\questions} \\
\midrule
\multicolumn{3}{@{}l}{\textbf{Sonnet~5}} \\
1 & $-0.07$ & $-0.26\;[-0.31,-0.20]$ \\
2 & $0.02$ & $0.28\;[0.23,0.34]$ \\
3 & $0.09$ & $0.40\;[0.35,0.44]$ \\
\midrule
\multicolumn{3}{@{}l}{\textbf{GPT-6.1~Sol}} \\
1 & $0.12$ & $-0.58\;[-0.68,-0.50]$ \\
2 & $0.08$ & $-0.33\;[-0.38,-0.28]$ \\
3 & $0.13$ & $-0.06\;[-0.11,-0.02]$ \\
\bottomrule
\end{tabular}
\end{center}

Question generation significantly improves Sonnet~5 at $B=2,3$:
the paired NG gains over predefined questions are $+0.27$
($95\%$ CI $[0.17,0.36]$) and $+0.31$ ($[0.25,0.37]$).
It instead reduces GPT-6.1~Sol's NG by $0.41$ ($[0.35,0.48]$)
and $0.20$ ($[0.14,0.25]$), respectively.
Differences are computed before rounding.
For context, generated-question VoI under Sonnet~5 at $B=3$ has no
significant difference from \method{} (VoI minus \method{}:
$-0.01$, $95\%$ CI $[-0.07,0.05]$).
Because the available questions differ, this comparison does not isolate
allocation quality under our fixed-candidate assumption.

A diagnostic that scores the same generated question trees with both
executors shows that their normalized gains depend on the executor:
Sol-generated trees reach NG $0.15/0.42$ with Sonnet~5 at $B=2/3$,
versus $-0.33/-0.06$ with Sol.
The normalization references are specific to each executor.
Sol's proposer also returns only one distinct candidate in $34\%$ of
turns, versus $9\%$ for Sonnet~5.
Thus, the benefit of question generation depends on both the questions
proposed and the executor using their answers.

\begin{table}[t]
\centering
\scriptsize
\setlength{\tabcolsep}{3.5pt}
\begin{tabular}{lccc}
\toprule
& \multicolumn{2}{c}{Batch} & Sequential \\
\cmidrule(lr){2-3}
Hyperparameter & \method{} & Local Value & Both \\
\midrule
Hidden width & 64 & 64 & 64 \\
Optimizer & AdamW & AdamW & AdamW \\
Learning rate & $3\times10^{-3}$ & $3\times10^{-3}$ & $3\times10^{-3}$ \\
Weight decay & $10^{-2}$ & $10^{-2}$ & $10^{-2}$ \\
Epochs & 400 & 400 & 300 \\
Batch size & Full batch & Full batch & 1024 \\
Dropout & 0 & 0.2 & 0.1 \\
Input standardization & No & Yes & Yes \\
Target scaling & $\div$ std. & $\div$ std. & $\div$ std. \\
Ensemble size & 5 & 5 & 5 \\
Training runs & 3 & 3 & 3 \\
Early stopping & No & No & No \\
\bottomrule
\end{tabular}
\caption{Training hyperparameters for \method{} and the local
value control baseline. Configurations are shared across
benchmarks and executors within each setting. Each training
run fits an ensemble of five independently initialized models,
whose predictions are averaged. Results are aggregated over
three independent training runs.}
\label{tab:app_hparams}
\end{table}

\begin{table}[t]
\centering
\scriptsize
\setlength{\tabcolsep}{2.5pt}
\begin{tabular}{llcrrr}
\toprule
Benchmark & Setting & Input dim. &
Train. examples & \method{} & Local Value \\
\midrule
Animals & Batch & $10+2$ & 15,680 & 9,281 & 897 \\
Trip & Batch & $8+2$ & 4,168 & 9,153 & 769 \\
Meeting & Batch & $10+6$ & 3,440 & 9,537 & 1,153 \\
Flight & Batch & $20+4$ & 2,938 & 10,049 & 1,665 \\
TB-D & Batch & $21+1$ & 360 & 9,921 & 1,537 \\
\midrule
Animals & Seq. & 28 & 251,034 & 6,081 & 6,081 \\
TB-D & Seq. & 146 & 412 / 1,140 & 13,698 & 13,633 \\
TB-H & Seq. & 129 & 8,904 / 8,234 & 12,610 & 12,545 \\
\bottomrule
\end{tabular}
\caption{Input dimensions, training examples, and trainable
parameters per model for \method{} and the local value
control baseline. Batch input dimensions are $d_z+d_c$,
and training examples are labeled consultation sets,
including the empty set. Sequential examples are decision
nodes; paired counts correspond to budgets $B=1/2$
(TB-D) or Sonnet~5/GPT-6.1~Sol executors (TB-H).
Each allocator ensembles five models, so total parameter
counts are five times those shown. The local value counts
exclude unused auxiliary output heads.}
\label{tab:app_params}
\end{table}

\subsection{Cardinality Control}
\label{appendix:cardinality}

Our batch formulation permits selecting fewer than the available $B$
consultations, whereas Local Value, Prompt, and VoI select exactly $B$.
This difference does not explain the gains of \method{}. On Animals, Trip
Planning, Meeting Planning, and TB-D, \method{} selects exactly $B$
consultations on every test instance under both executors and every evaluated
budget, so constraining it to $|S|=B$ leaves all results unchanged.

Flight Recommendation is the only exception. At $B=3$, \method{} selects fewer
than three consultations on 72\% of Sonnet~5 instances and 67\% of GPT-6.1
Sol instances. This stopping freedom lowers performance: forcing exactly
three consultations increases normalized gain from $0.33$ to $0.39$ for
Sonnet~5 and from $-0.28$ to $0.31$ for GPT-6.1 Sol. Thus, the reported
$|S|\le B$ protocol is conservative on the only benchmark where cardinality
freedom is exercised. See Table~\ref{tab:cardinality_control} for the comparison.

\begin{table}[t]
\centering
\small
\setlength{\tabcolsep}{1pt}
\begin{tabular}{lccc}
\toprule
Executor & $\Pr(|S|<B)$ & \method{} ($|S|\le B$) & \method{} ($|S|=B$) \\
\midrule
Sonnet~5      & 72\% & 0.33  & 0.39 \\
GPT-6.1 Sol   & 67\% & -0.28 & 0.31 \\
\bottomrule
\end{tabular}
\caption{Cardinality control on Flight Recommendation at $B=3$.
Entries are unclipped normalized gain. Constraining \method{} to use the full budget
improves performance. All other batch benchmarks already use exactly $B$ on
every test instance.}
\label{tab:cardinality_control}
\end{table}

Allowing Local Value to select only positive-valued consultations changes
results only on Flight and yields mixed effects, providing no systematic
advantage from matching the stopping rule.

\section{Full Main Results}
\label{appendix:full_results}

Tables~\ref{tab:app_batch_util}--\ref{tab:app_seq_ng} report the complete frozen test results for all evaluated budgets. For both batch and sequential settings, we report raw downstream utility and unclipped normalized gain. The normalized gain tables additionally include $95\%$ bootstrap intervals over test episodes.
In these tables,
\begin{itemize}[leftmargin=*, topsep=0mm, itemsep=0mm]
    \item \emph{None}: no consultation.
    \item \emph{Random}: random allocation under the available budget.
    \item \emph{Prompt}: direct LLM allocation averaged over three samples.
    \item \emph{VoI}: the released implementation of \cite{dong2026voi} where applicable.
    \item \emph{Local Value}: our independent local-value control.
    \item \emph{\method{}}: the proposed total-value predictor.
    \item \emph{Oracle}: the normalization reference defined in Appendix~\ref{appendix:metrics_eval}.
\end{itemize}

In the sequential tables,
\emph{Batch set} applies the batch \method{} selected set without adaptation
(Animals), while \emph{Best fixed set} is the privileged fixed set selected
using label-half values. For sequential Animals, normalized gain uses
Random as its zero reference. And for TB-D and TB-H, it uses no consultation.

\paragraph{Meeting Planning sensitivity.}
Meeting Planning contains only 20 test tasks, making its method comparisons the least precise in the suite. The \method{} and Local Value comparison at $B=4$ is also sensitive to whether the 20 development instances are included in final training. Using only the 60 training instances, \method{} exceeds Local by $+0.15$ under Sonnet~5 and $+0.02$ under GPT-6.1~Sol. Under the final
protocol, which trains on all 80 non-test instances, the differences become $+0.09$ and $-0.06$, respectively, and neither is significant. We therefore treat Meeting Planning as showing no reliable difference between \method{} and Local Value.

\begin{table*}[t]
\centering
\scriptsize
\setlength{\tabcolsep}{1.5pt}
\begin{tabular}{lllrccccccc}
\toprule
& Benchmark & $B$ & $n_{\rm test}$ & None & Random & Prompt & VoI & \makecell{Local Value\\(ours)} & \makecell{\method{}\\(ours)} & Oracle \\
\midrule
\multirow{12}{*}{\rotatebox{90}{Sonnet 5}}
& Animals & 1 & 75 & $-$0.9800 & $-$0.9556 & $-$0.9537\,{\tiny$\pm$0.0008} & $-$0.9565\,{\tiny$\pm$0.0002} & $-$0.9459\,{\tiny$\pm$0.0000} & $-$0.9476\,{\tiny$\pm$0.0001} & $-$0.9414 \\
&  & 2 &  & $-$0.9800 & $-$0.9198 & $-$0.9217\,{\tiny$\pm$0.0012} & $-$0.9223\,{\tiny$\pm$0.0004} & $-$0.9030\,{\tiny$\pm$0.0005} & $-$0.9037\,{\tiny$\pm$0.0006} & $-$0.8921 \\
&  & 3 &  & $-$0.9800 & $-$0.8716 & $-$0.8655\,{\tiny$\pm$0.0005} & $-$0.8697\,{\tiny$\pm$0.0013} & $-$0.8622\,{\tiny$\pm$0.0008} & $-$0.8499\,{\tiny$\pm$0.0006} & $-$0.8343 \\
& Trip & 1 & 79 & 0.027 & 0.135 & 0.129\,{\tiny$\pm$0.002} & --- & 0.174\,{\tiny$\pm$0.001} & 0.173\,{\tiny$\pm$0.001} & 0.176 \\
&  & 2 &  & 0.027 & 0.363 & 0.370\,{\tiny$\pm$0.004} & --- & 0.426\,{\tiny$\pm$0.006} & 0.429\,{\tiny$\pm$0.001} & 0.443 \\
& Meeting & 1 & 20 & 2.170 & 2.284 & 2.361\,{\tiny$\pm$0.026} & --- & 2.595\,{\tiny$\pm$0.003} & 2.619\,{\tiny$\pm$0.012} & 2.634 \\
&  & 2 &  & 2.170 & 2.438 & 2.418\,{\tiny$\pm$0.050} & --- & 3.001\,{\tiny$\pm$0.014} & 3.003\,{\tiny$\pm$0.012} & 3.108 \\
&  & 4 &  & 2.170 & 2.804 & 2.768\,{\tiny$\pm$0.034} & --- & 3.878\,{\tiny$\pm$0.024} & 3.979\,{\tiny$\pm$0.043} & 3.982 \\
& Flight & 1 & 30 & 0.574 & 0.631 & 0.619\,{\tiny$\pm$0.004} & 0.641\,{\tiny$\pm$0.012} & 0.698\,{\tiny$\pm$0.023} & 0.702\,{\tiny$\pm$0.006} & 0.811 \\
&  & 2 &  & 0.574 & 0.673 & 0.701\,{\tiny$\pm$0.010} & 0.717\,{\tiny$\pm$0.010} & 0.729\,{\tiny$\pm$0.006} & 0.740\,{\tiny$\pm$0.003} & 0.816 \\
&  & 3 &  & 0.574 & 0.715 & 0.748\,{\tiny$\pm$0.006} & 0.740\,{\tiny$\pm$0.013} & 0.711\,{\tiny$\pm$0.004} & 0.760\,{\tiny$\pm$0.005} & 0.849 \\
& TB-D & 1 & 60 & 0.716 & 1.783 & 2.560\,{\tiny$\pm$0.080} & --- & 3.165\,{\tiny$\pm$0.003} & 3.147\,{\tiny$\pm$0.013} & 3.126 \\
\midrule
\multirow{12}{*}{\rotatebox{90}{GPT-6.1 Sol}}
& Animals & 1 & 75 & $-$0.9790 & $-$0.9490 & $-$0.9461\,{\tiny$\pm$0.0005} & $-$0.9469\,{\tiny$\pm$0.0001} & $-$0.9395\,{\tiny$\pm$0.0001} & $-$0.9411\,{\tiny$\pm$0.0001} & $-$0.9313 \\
&  & 2 &  & $-$0.9790 & $-$0.8859 & $-$0.8796\,{\tiny$\pm$0.0004} & $-$0.8838\,{\tiny$\pm$0.0017} & $-$0.8689\,{\tiny$\pm$0.0000} & $-$0.8736\,{\tiny$\pm$0.0019} & $-$0.8431 \\
&  & 3 &  & $-$0.9790 & $-$0.7885 & $-$0.7769\,{\tiny$\pm$0.0020} & $-$0.7782\,{\tiny$\pm$0.0015} & $-$0.7830\,{\tiny$\pm$0.0003} & $-$0.7643\,{\tiny$\pm$0.0014} & $-$0.7281 \\
& Trip & 1 & 79 & 0.035 & 0.156 & 0.191\,{\tiny$\pm$0.005} & --- & 0.213\,{\tiny$\pm$0.000} & 0.207\,{\tiny$\pm$0.001} & 0.215 \\
&  & 2 &  & 0.035 & 0.397 & 0.422\,{\tiny$\pm$0.010} & --- & 0.458\,{\tiny$\pm$0.001} & 0.462\,{\tiny$\pm$0.004} & 0.478 \\
& Meeting & 1 & 20 & 2.114 & 2.275 & 2.224\,{\tiny$\pm$0.005} & --- & 2.642\,{\tiny$\pm$0.006} & 2.610\,{\tiny$\pm$0.004} & 2.735 \\
&  & 2 &  & 2.114 & 2.456 & 2.555\,{\tiny$\pm$0.079} & --- & 3.208\,{\tiny$\pm$0.010} & 3.154\,{\tiny$\pm$0.017} & 3.364 \\
&  & 4 &  & 2.114 & 2.941 & 3.342\,{\tiny$\pm$0.051} & --- & 4.521\,{\tiny$\pm$0.033} & 4.419\,{\tiny$\pm$0.028} & 4.735 \\
& Flight & 1 & 30 & 0.728 & 0.738 & 0.755\,{\tiny$\pm$0.003} & 0.758\,{\tiny$\pm$0.004} & 0.728\,{\tiny$\pm$0.012} & 0.728\,{\tiny$\pm$0.001} & 0.823 \\
&  & 2 &  & 0.728 & 0.756 & 0.772\,{\tiny$\pm$0.003} & 0.771\,{\tiny$\pm$0.007} & 0.754\,{\tiny$\pm$0.007} & 0.746\,{\tiny$\pm$0.020} & 0.850 \\
&  & 3 &  & 0.728 & 0.782 & 0.804\,{\tiny$\pm$0.001} & 0.793\,{\tiny$\pm$0.006} & 0.789\,{\tiny$\pm$0.008} & 0.761\,{\tiny$\pm$0.018} & 0.858 \\
& TB-D & 1 & 60 & 0.830 & 1.864 & 2.608\,{\tiny$\pm$0.030} & --- & 3.237\,{\tiny$\pm$0.014} & 3.202\,{\tiny$\pm$0.001} & 3.238 \\
\bottomrule
\end{tabular}
\caption{
    \textbf{Batch setting: raw test utility for all benchmarks, budgets, and executors.} Utilities are reported in native units: Animals, $2\Pr(\mathrm{correct})-1$; Trip Planning, fraction of valid itineraries; Meeting Planning, meetings achieved; Flight, per-task normalized preference utility; and TB-D, points. Local Value and \method{} report mean $\pm$ SD across three independent training runs; Prompt and VoI report mean $\pm$ SD across three independent stochastic test runs. None, Random, and Oracle are fixed given the evaluation labels. These SDs reflect run-to-run variation. Uncertainty over test episodes is reported in the normalized-gain tables.
}
\label{tab:app_batch_util}
\end{table*}

\begin{table*}[t]
\centering
\scriptsize
\setlength{\tabcolsep}{1.5pt}
\begin{tabular}{lllrccccccc}
\toprule
& Benchmark & $B$ & $n_{\rm test}$ & None & Random & Prompt & VoI & \makecell{Local Value\\(ours)} & \makecell{\method{}\\(ours)} & Oracle \\
\midrule
\multirow{12}{*}{\rotatebox{90}{Sonnet 5}}
& Animals & 1 & 75 & $-$1.72\,{\tiny[$-$2.13,$-$1.41]} & 0.00 & 0.13\,{\tiny[0.02,0.26]} & $-$0.07\,{\tiny[$-$0.17,0.06]} & 0.68\,{\tiny[0.51,0.82]} & 0.56\,{\tiny[0.39,0.75]} & 1.00 \\
&  & 2 &  & $-$2.17\,{\tiny[$-$2.64,$-$1.82]} & 0.00 & $-$0.07\,{\tiny[$-$0.17,0.03]} & $-$0.09\,{\tiny[$-$0.18,0.01]} & 0.61\,{\tiny[0.44,0.76]} & 0.58\,{\tiny[0.42,0.72]} & 1.00 \\
&  & 3 &  & $-$2.90\,{\tiny[$-$3.43,$-$2.53]} & 0.00 & 0.16\,{\tiny[0.01,0.32]} & 0.05\,{\tiny[$-$0.04,0.14]} & 0.25\,{\tiny[0.10,0.40]} & 0.58\,{\tiny[0.42,0.81]} & 1.00 \\
& Trip & 1 & 79 & $-$2.64\,{\tiny[$-$4.48,$-$1.90]} & 0.00 & $-$0.16\,{\tiny[$-$0.56,0.27]} & --- & 0.95\,{\tiny[0.73,1.17]} & 0.92\,{\tiny[0.69,1.15]} & 1.00 \\
&  & 2 &  & $-$4.18\,{\tiny[$-$5.88,$-$3.21]} & 0.00 & 0.09\,{\tiny[$-$0.24,0.41]} & --- & 0.79\,{\tiny[0.54,1.06]} & 0.83\,{\tiny[0.59,1.08]} & 1.00 \\
& Meeting & 1 & 20 & $-$0.33\,{\tiny[$-$0.47,$-$0.24]} & 0.00 & 0.22\,{\tiny[$-$0.09,0.54]} & --- & 0.89\,{\tiny[0.67,1.08]} & 0.96\,{\tiny[0.73,1.18]} & 1.00 \\
&  & 2 &  & $-$0.40\,{\tiny[$-$0.47,$-$0.34]} & 0.00 & $-$0.03\,{\tiny[$-$0.18,0.12]} & --- & 0.84\,{\tiny[0.64,1.04]} & 0.84\,{\tiny[0.65,1.06]} & 1.00 \\
&  & 4 &  & $-$0.54\,{\tiny[$-$0.61,$-$0.48]} & 0.00 & $-$0.03\,{\tiny[$-$0.18,0.12]} & --- & 0.91\,{\tiny[0.83,1.01]} & 1.00\,{\tiny[0.90,1.12]} & 1.00 \\
& Flight & 1 & 30 & $-$0.31\,{\tiny[$-$0.67,0.00]} & 0.00 & $-$0.07\,{\tiny[$-$0.31,0.17]} & 0.06\,{\tiny[$-$0.17,0.30]} & 0.37\,{\tiny[0.04,0.67]} & 0.40\,{\tiny[$-$0.01,0.70]} & 1.00 \\
&  & 2 &  & $-$0.69\,{\tiny[$-$1.35,$-$0.21]} & 0.00 & 0.20\,{\tiny[$-$0.05,0.47]} & 0.31\,{\tiny[$-$0.01,0.64]} & 0.39\,{\tiny[0.07,0.67]} & 0.47\,{\tiny[$-$0.01,0.76]} & 1.00 \\
&  & 3 &  & $-$1.05\,{\tiny[$-$1.70,$-$0.48]} & 0.00 & 0.24\,{\tiny[0.05,0.46]} & 0.18\,{\tiny[$-$0.10,0.48]} & $-$0.03\,{\tiny[$-$0.33,0.26]} & 0.33\,{\tiny[$-$0.07,0.64]} & 1.00 \\
& TB-D & 1 & 60 & $-$0.79\,{\tiny[$-$0.92,$-$0.69]} & 0.00 & 0.58\,{\tiny[0.43,0.72]} & --- & 1.03\,{\tiny[0.97,1.10]} & 1.02\,{\tiny[0.96,1.08]} & 1.00 \\
\midrule
\multirow{12}{*}{\rotatebox{90}{GPT-6.1 Sol}}
& Animals & 1 & 75 & $-$1.70\,{\tiny[$-$2.06,$-$1.42]} & 0.00 & 0.16\,{\tiny[0.03,0.30]} & 0.12\,{\tiny[0.02,0.24]} & 0.54\,{\tiny[0.36,0.69]} & 0.45\,{\tiny[0.27,0.61]} & 1.00 \\
&  & 2 &  & $-$2.17\,{\tiny[$-$2.47,$-$1.93]} & 0.00 & 0.15\,{\tiny[0.05,0.25]} & 0.05\,{\tiny[$-$0.03,0.14]} & 0.40\,{\tiny[0.26,0.54]} & 0.29\,{\tiny[0.16,0.41]} & 1.00 \\
&  & 3 &  & $-$3.16\,{\tiny[$-$3.49,$-$2.87]} & 0.00 & 0.19\,{\tiny[0.10,0.29]} & 0.17\,{\tiny[0.07,0.27]} & 0.09\,{\tiny[$-$0.02,0.21]} & 0.40\,{\tiny[0.30,0.50]} & 1.00 \\
& Trip & 1 & 79 & $-$2.07\,{\tiny[$-$2.84,$-$1.63]} & 0.00 & 0.60\,{\tiny[0.36,0.78]} & --- & 0.97\,{\tiny[0.88,1.05]} & 0.86\,{\tiny[0.71,0.98]} & 1.00 \\
&  & 2 &  & $-$4.46\,{\tiny[$-$5.92,$-$3.58]} & 0.00 & 0.31\,{\tiny[0.04,0.54]} & --- & 0.75\,{\tiny[0.52,0.97]} & 0.80\,{\tiny[0.56,1.04]} & 1.00 \\
& Meeting & 1 & 20 & $-$0.35\,{\tiny[$-$0.46,$-$0.27]} & 0.00 & $-$0.11\,{\tiny[$-$0.23,0.06]} & --- & 0.80\,{\tiny[0.55,1.00]} & 0.73\,{\tiny[0.46,0.96]} & 1.00 \\
&  & 2 &  & $-$0.38\,{\tiny[$-$0.43,$-$0.33]} & 0.00 & 0.11\,{\tiny[$-$0.06,0.28]} & --- & 0.83\,{\tiny[0.71,0.94]} & 0.77\,{\tiny[0.64,0.89]} & 1.00 \\
&  & 4 &  & $-$0.46\,{\tiny[$-$0.51,$-$0.41]} & 0.00 & 0.22\,{\tiny[0.10,0.36]} & --- & 0.88\,{\tiny[0.75,0.99]} & 0.82\,{\tiny[0.73,0.92]} & 1.00 \\
& Flight & 1 & 30 & $-$0.11\,{\tiny[$-$0.79,0.27]} & 0.00 & 0.21\,{\tiny[$-$0.12,0.55]} & 0.24\,{\tiny[$-$0.12,0.71]} & $-$0.11\,{\tiny[$-$0.58,0.36]} & $-$0.12\,{\tiny[$-$0.64,0.29]} & 1.00 \\
&  & 2 &  & $-$0.29\,{\tiny[$-$0.95,0.17]} & 0.00 & 0.17\,{\tiny[$-$0.23,0.62]} & 0.16\,{\tiny[$-$0.21,0.49]} & $-$0.02\,{\tiny[$-$0.47,0.47]} & $-$0.11\,{\tiny[$-$0.58,0.32]} & 1.00 \\
&  & 3 &  & $-$0.71\,{\tiny[$-$1.97,$-$0.04]} & 0.00 & 0.29\,{\tiny[$-$0.05,0.82]} & 0.14\,{\tiny[$-$0.21,0.43]} & 0.09\,{\tiny[$-$0.25,0.44]} & $-$0.28\,{\tiny[$-$0.98,0.22]} & 1.00 \\
& TB-D & 1 & 60 & $-$0.75\,{\tiny[$-$0.86,$-$0.67]} & 0.00 & 0.54\,{\tiny[0.35,0.70]} & --- & 1.00\,{\tiny[0.95,1.05]} & 0.97\,{\tiny[0.91,1.04]} & 1.00 \\
\bottomrule
\end{tabular}
\caption{\textbf{Batch setting: unclipped normalized gain for all benchmarks, budgets, and executors}, with \(95\%\) bootstrap intervals over test episodes. Random and Oracle equal 0 and 1, respectively, by normalization; values outside $[0,1]$ are retained.}
\label{tab:app_batch_ng}
\end{table*}

\begin{table*}[t]
\centering
\scriptsize
\setlength{\tabcolsep}{1.5pt}
\begin{tabular}{lllrccccccccc}
\toprule
& Benchmark & $B$ & $n_{\rm test}$ & None & Random & Prompt & VoI & \makecell{Local Value\\(ours)} & \makecell{\method{}\\(ours)} & \makecell{Batch\\set} & \makecell{Best\\fixed set} & Oracle \\
\midrule
\multirow{5}{*}{\rotatebox{90}{Sonnet 5}}
& Animals & 2 & 75 & $-$0.9800 & $-$0.9198 & $-$0.9185\,{\tiny$\pm$0.0003} & $-$0.9191\,{\tiny$\pm$0.0023} & $-$0.8928\,{\tiny$\pm$0.0001} & $-$0.8958\,{\tiny$\pm$0.0020} & $-$0.9037 & $-$0.8921 & $-$0.8760 \\
&  & 3 &  & $-$0.9800 & $-$0.8716 & $-$0.8376\,{\tiny$\pm$0.0014} & $-$0.8626\,{\tiny$\pm$0.0032} & $-$0.8286\,{\tiny$\pm$0.0005} & $-$0.8307\,{\tiny$\pm$0.0029} & $-$0.8499 & $-$0.8337 & $-$0.7698 \\
& TB-D & 1 & 60 & 0.716 & --- & 2.571\,{\tiny$\pm$0.049} & --- & 2.923\,{\tiny$\pm$0.026} & 3.101\,{\tiny$\pm$0.004} & --- & 3.150 & 3.250 \\
&  & 2 &  & 0.716 & --- & 3.643\,{\tiny$\pm$0.014} & --- & 4.099\,{\tiny$\pm$0.029} & 4.412\,{\tiny$\pm$0.010} & --- & 4.370 & 4.607 \\
& TB-H & 2 & 30 & $-$1.483 & --- & 1.681\,{\tiny$\pm$0.075} & --- & 3.467\,{\tiny$\pm$0.041} & 4.169\,{\tiny$\pm$0.020} & --- & 4.075 & 4.742 \\
\midrule
\multirow{5}{*}{\rotatebox{90}{GPT-6.1 Sol}}
& Animals & 2 & 75 & $-$0.9790 & $-$0.8859 & $-$0.8793\,{\tiny$\pm$0.0040} & $-$0.8808\,{\tiny$\pm$0.0020} & $-$0.8576\,{\tiny$\pm$0.0007} & $-$0.8600\,{\tiny$\pm$0.0016} & $-$0.8736 & $-$0.8433 & $-$0.8255 \\
&  & 3 &  & $-$0.9790 & $-$0.7885 & $-$0.7641\,{\tiny$\pm$0.0026} & $-$0.7709\,{\tiny$\pm$0.0011} & $-$0.7341\,{\tiny$\pm$0.0011} & $-$0.7409\,{\tiny$\pm$0.0020} & $-$0.7643 & $-$0.7277 & $-$0.6578 \\
& TB-D & 1 & 60 & 0.830 & --- & 2.812\,{\tiny$\pm$0.024} & --- & 2.982\,{\tiny$\pm$0.009} & 3.149\,{\tiny$\pm$0.017} & --- & 3.267 & 3.326 \\
&  & 2 &  & 0.830 & --- & 3.788\,{\tiny$\pm$0.027} & --- & 4.159\,{\tiny$\pm$0.000} & 4.509\,{\tiny$\pm$0.013} & --- & 4.453 & 4.661 \\
& TB-H & 2 & 30 & $-$1.717 & --- & 1.583\,{\tiny$\pm$0.000} & --- & 3.156\,{\tiny$\pm$0.167} & 3.842\,{\tiny$\pm$0.035} & --- & 3.867 & 4.383 \\
\bottomrule
\end{tabular}
\caption{
    \textbf{Sequential setting: raw test utility for all benchmarks, budgets, and executors.} Utilities are reported in native units: Animals, $2\Pr(\mathrm{correct})-1$; TB-D and TB-H, points. Local Value and CVD report mean $\pm$ SD across three independent training runs; Prompt and VoI report mean $\pm$ SD across three independent stochastic test runs. None, Random, and other reference allocators are fixed given the evaluation labels. These SDs reflect run-to-run variation. Uncertainty over test episodes is reported in the normalized-gain tables.
}
\label{tab:app_seq_util}
\end{table*}

\begin{table*}[t]
\centering
\scriptsize
\setlength{\tabcolsep}{1.5pt}
\begin{tabular}{lllrccccccccc}
\toprule
& Benchmark & $B$ & $n_{\rm test}$ & None & Random & Prompt & VoI & \makecell{Local Value\\(ours)} & \makecell{\method{}\\(ours)} & \makecell{Batch\\set} & \makecell{Best\\fixed set} & Oracle \\
\midrule
\multirow{5}{*}{\rotatebox{90}{Sonnet 5}}
& Animals & 2 & 75 & $-$1.37\,{\tiny[$-$1.55,$-$1.22]} & 0.00 & 0.03\,{\tiny[$-$0.01,0.07]} & 0.02\,{\tiny[$-$0.05,0.08]} & 0.62\,{\tiny[0.52,0.71]} & 0.55\,{\tiny[0.45,0.64]} & 0.37\,{\tiny[0.25,0.47]} & 0.63\,{\tiny[0.56,0.70]} & 1.00 \\
&  & 3 &  & $-$1.06\,{\tiny[$-$1.14,$-$0.99]} & 0.00 & 0.33\,{\tiny[0.29,0.38]} & 0.09\,{\tiny[0.05,0.13]} & 0.42\,{\tiny[0.36,0.48]} & 0.40\,{\tiny[0.35,0.46]} & 0.21\,{\tiny[0.16,0.28]} & 0.37\,{\tiny[0.33,0.42]} & 1.00 \\
& TB-D & 1 & 60 & 0.00 & --- & 0.73\,{\tiny[0.66,0.80]} & --- & 0.87\,{\tiny[0.81,0.92]} & 0.94\,{\tiny[0.90,0.97]} & --- & 0.96\,{\tiny[0.93,0.99]} & 1.00 \\
&  & 2 &  & 0.00 & --- & 0.75\,{\tiny[0.70,0.80]} & --- & 0.87\,{\tiny[0.83,0.91]} & 0.95\,{\tiny[0.92,0.97]} & --- & 0.94\,{\tiny[0.92,0.96]} & 1.00 \\
& TB-H & 2 & 30 & 0.00 & --- & 0.51\,{\tiny[0.42,0.59]} & --- & 0.80\,{\tiny[0.73,0.86]} & 0.91\,{\tiny[0.88,0.94]} & --- & 0.89\,{\tiny[0.85,0.93]} & 1.00 \\
\midrule
\multirow{5}{*}{\rotatebox{90}{GPT-6.1 Sol}}
& Animals & 2 & 75 & $-$1.54\,{\tiny[$-$1.70,$-$1.40]} & 0.00 & 0.11\,{\tiny[0.04,0.19]} & 0.08\,{\tiny[0.03,0.14]} & 0.47\,{\tiny[0.37,0.56]} & 0.43\,{\tiny[0.35,0.51]} & 0.20\,{\tiny[0.11,0.29]} & 0.71\,{\tiny[0.65,0.77]} & 1.00 \\
&  & 3 &  & $-$1.46\,{\tiny[$-$1.55,$-$1.38]} & 0.00 & 0.19\,{\tiny[0.13,0.24]} & 0.13\,{\tiny[0.09,0.18]} & 0.42\,{\tiny[0.35,0.48]} & 0.36\,{\tiny[0.31,0.41]} & 0.19\,{\tiny[0.14,0.23]} & 0.46\,{\tiny[0.43,0.50]} & 1.00 \\
& TB-D & 1 & 60 & 0.00 & --- & 0.79\,{\tiny[0.73,0.86]} & --- & 0.86\,{\tiny[0.79,0.92]} & 0.93\,{\tiny[0.89,0.97]} & --- & 0.98\,{\tiny[0.95,0.99]} & 1.00 \\
&  & 2 &  & 0.00 & --- & 0.77\,{\tiny[0.72,0.83]} & --- & 0.87\,{\tiny[0.82,0.91]} & 0.96\,{\tiny[0.93,0.98]} & --- & 0.95\,{\tiny[0.92,0.97]} & 1.00 \\
& TB-H & 2 & 30 & 0.00 & --- & 0.54\,{\tiny[0.47,0.61]} & --- & 0.80\,{\tiny[0.75,0.84]} & 0.91\,{\tiny[0.87,0.95]} & --- & 0.92\,{\tiny[0.87,0.96]} & 1.00 \\
\bottomrule
\end{tabular}
\caption{
    \textbf{Sequential setting: unclipped normalized gain for all benchmarks, budgets, and executors}, with $95\%$ bootstrap intervals over test episodes. For Animals, Random defines 0; for TB-D and TB-H, None defines 0; Oracle defines 1 in all cases. Values outside $[0,1]$ are retained.
}
\label{tab:app_seq_ng}
\end{table*}

\subsection{Extended TB-H Coupled/Control Evaluation}
\label{appendix:tbh_coupling}

To examine whether state coupling increases the benefit of
continuation-aware allocation, we add 36 matched coupled/control pairs
from the original TB-H generator: 12 per workflow family, using seeds
$30000$--$30011$. Together with the original 15 pairs, this gives 51
pairs (102 worlds). Within a pair, task content is matched; the coupled
variant's later conditions depend on earlier decisions, whereas the
control uses independent public randomness for the corresponding
triggers. Both variants retain the shared consultation budget $B=2$.

We use the original labeling and evaluation protocol, with eight hidden
completions per world split into four label-half and four evaluation-half
completions. Policies use the original training data and three training
seeds, with no fitting on the new worlds. Evaluation uses the same
approximate replay described in Appendix~\ref{appendix:seq_targets}.
Construction checks confirm matched task content and the absence of
state-mediated dependencies in the controls; frozen-policy checks
reproduce the original test results.

The table reports \method{} minus Local Value in raw utility units
(points), averaged over the three training runs. The final column is
the paired difference between these advantages: coupled minus control.
Brackets give $95\%$ bootstrap intervals, resampling matched pairs
together. Differences are computed before rounding. Original, new,
and pooled results are shown separately; the preceding main-result
tables retain the original test set.

\begin{center}
\footnotesize
\setlength{\tabcolsep}{2pt}
\begin{tabular}{@{}lccc@{}}
\toprule
Pairs & Coupled & Control & \makecell{Coupled\\minus control} \\
\midrule
\multicolumn{4}{@{}l}{\textbf{Sonnet~5}} \\
Original 15 &
\makecell{$+0.28$\\{\scriptsize$[-0.13,0.78]$}} &
\makecell{$+1.13$\\{\scriptsize$[0.51,1.79]$}} &
\makecell{$-0.85$\\{\scriptsize$[-1.67,0.00]$}} \\
New 36 &
\makecell{$+0.50$\\{\scriptsize$[0.16,0.86]$}} &
\makecell{$+0.99$\\{\scriptsize$[0.55,1.44]$}} &
\makecell{$-0.49$\\{\scriptsize$[-0.97,0.01]$}} \\
Pooled 51 &
\makecell{$+0.43$\\{\scriptsize$[0.16,0.73]$}} &
\makecell{$+1.03$\\{\scriptsize$[0.67,1.41]$}} &
\makecell{$-0.59$\\{\scriptsize$[-1.00,-0.19]$}} \\
\midrule
\multicolumn{4}{@{}l}{\textbf{GPT-6.1~Sol}} \\
Original 15 &
\makecell{$+0.54$\\{\scriptsize$[-0.20,1.21]$}} &
\makecell{$+0.83$\\{\scriptsize$[0.54,1.09]$}} &
\makecell{$-0.29$\\{\scriptsize$[-0.87,0.29]$}} \\
New 36 &
\makecell{$+0.71$\\{\scriptsize$[0.32,1.08]$}} &
\makecell{$+0.57$\\{\scriptsize$[0.28,0.84]$}} &
\makecell{$+0.14$\\{\scriptsize$[-0.25,0.53]$}} \\
Pooled 51 &
\makecell{$+0.66$\\{\scriptsize$[0.33,1.00]$}} &
\makecell{$+0.65$\\{\scriptsize$[0.44,0.86]$}} &
\makecell{$+0.01$\\{\scriptsize$[-0.31,0.33]$}} \\
\bottomrule
\end{tabular}
\end{center}

On the new pairs alone and in the pooled evaluation, \method{}
significantly outperforms Local Value in both variants under both
executors. In the pooled evaluation, its advantage is smaller under
coupling with Sonnet~5 and does not differ significantly with
GPT-6.1~Sol. These results show that the benefit of continuation-aware
allocation is not confined to state-coupled workflows. Even in the
control variant, consulting now consumes capacity that may be more
valuable at a later decision. This finding does not imply that state
coupling is irrelevant to allocation.

\section{Method Ablation Details}
\label{appendix:add_results}

\paragraph{Compared variants.}
We compare direct total-value regression with several natural alternatives.
\begin{enumerate}[leftmargin=*, topsep=0mm, itemsep=0mm]
    \item \emph{Explicit decompositions} represent set value using per-consultation
    values together with an interaction component: a residual over a frozen
    local-value predictor (R), a protected residual that preserves singleton
    predictions (PR), a pairwise residual (S2), and jointly trained additive or
    local+interaction models.

    \item \emph{Local-value auxiliaries} provide learned local values as predictor
inputs (D+v), add an auxiliary local-value loss, or increase predictor width.
    \item \emph{Relative supervision} includes a within-episode local-value learner,
Contrast-D with a pairwise ranking loss over consultation sets, and a hybrid
absolute+relative objective.

    \item \emph{Independent local-value selection and a single
predictor} without ensembling.
\end{enumerate}

All variants use the same data splits and base task features as \method{};
some additionally use learned local-value estimates as specified above.
Sequential variants are trained on the corresponding continuation-aware
or myopic targets.

\paragraph{Within-episode local learner.}
This learner predicts relative consultation values within each task.
It augments candidate features with learned local-value estimates, then
subtracts each task's mean from these inputs and its singleton-value labels.
A ridge regression model is trained on the centered quantities, with
cross-fitted local estimates for training tasks. At deployment, it ranks
consultations using centered inputs and selects the top $B$, without
observing the task's true consultation values or fitting a new model.
This is the Within-ep.\ column in Table~\ref{tab:app_ablation_b}.

\paragraph{Selection and evaluation protocol.}
Alternative families were screened on development data, with configuration
selection and hyperparameters fixed before evaluation on the frozen test
sets. Tables~\ref{tab:app_ablation}--\ref{tab:app_ablation_b} report the
selected configurations under both executors, using the main protocol:
five-model ensembles for the alternative and its direct-regression
reference, with per-episode utilities averaged over three training runs.
Each ensemble selects an allocation from its mean predictions.
The Single model ablation intentionally removes ensembling and averages
the utilities of the 15 individual predictors against the three-run
ensemble reference. Local Value retains the main-result protocol.
Meeting uses the direct-regression reference from its matched ablation
implementation; other references follow the corresponding learning
pipeline. Paired $95\%$ bootstrap intervals use 4,000 episode resamples,
with normalization references resampled jointly. Significance is determined
before rounding. The two tables contain 147 alternative--benchmark--budget
comparisons per executor.

\providecommand{\ablcell}[2]{\begin{tabular}[c]{@{}c@{}}#1\\[-2pt]{\scriptsize #2}\end{tabular}}
\providecommand{\ablmaincell}[2]{\begingroup\renewcommand{\arraystretch}{0.9}\begin{tabular}[c]{@{}c@{}}#1\\[-1.5pt]{\fontsize{7}{8}\selectfont #2}\end{tabular}\endgroup}
\begin{table*}[p]
\centering
\scriptsize
\setlength{\tabcolsep}{2pt}
\renewcommand{\arraystretch}{1.0}
\begin{tabular}{llcccccccc}
\toprule
\multicolumn{10}{c}{\textbf{Sonnet~5}} \\
& & \multicolumn{5}{c}{\textit{Explicit decomposition}} & \multicolumn{3}{c}{\textit{Local-value auxiliaries}} \\
\cmidrule(lr){3-7}\cmidrule(lr){8-10}
Bench. & $B$ & R & PR & S2 & Joint add. & Joint v+M & D+v & Aux.\ loss & Wider \\
\midrule
Animals & 1 & \ablmaincell{+0.09}{[$-$0.04, +0.27]} & \ablmaincell{+0.12}{[$-$0.00, +0.30]} & \ablmaincell{+0.12}{[$-$0.00, +0.30]} & \ablmaincell{+0.05}{[$-$0.03, +0.16]} & \ablmaincell{+0.05}{[$-$0.03, +0.16]} & \ablmaincell{+0.00}{[+0.00, +0.00]} & --- & --- \\
Animals & 2 & \ablmaincell{+0.02}{[$-$0.04, +0.09]} & \ablmaincell{$-$0.05$^{\ast}$}{[$-$0.09, $-$0.01]} & \ablmaincell{+0.03}{[$-$0.11, +0.16]} & \ablmaincell{+0.03}{[$-$0.05, +0.13]} & \ablmaincell{$-$0.08}{[$-$0.21, +0.04]} & \ablmaincell{$-$0.07$^{\ast}$}{[$-$0.15, $-$0.01]} & --- & --- \\
Animals & 3 & \ablmaincell{+0.08}{[$-$0.03, +0.19]} & \ablmaincell{+0.07}{[$-$0.01, +0.17]} & \ablmaincell{+0.04}{[$-$0.09, +0.18]} & \ablmaincell{$-$0.14$^{\ast}$}{[$-$0.26, $-$0.02]} & \ablmaincell{+0.05}{[$-$0.05, +0.16]} & \ablmaincell{+0.04}{[$-$0.02, +0.10]} & --- & --- \\
Trip & 1 & \ablmaincell{+0.07}{[$-$0.12, +0.30]} & \ablmaincell{+0.02}{[$-$0.20, +0.25]} & \ablmaincell{+0.02}{[$-$0.20, +0.25]} & \ablmaincell{+0.04}{[$-$0.14, +0.25]} & \ablmaincell{$-$0.04}{[$-$0.27, +0.20]} & \ablmaincell{+0.07$^{\ast}$}{[+0.00, +0.21]} & --- & --- \\
Trip & 2 & \ablmaincell{$-$0.00}{[$-$0.13, +0.13]} & \ablmaincell{+0.06}{[$-$0.08, +0.24]} & \ablmaincell{$-$0.12}{[$-$0.36, +0.08]} & \ablmaincell{$-$0.10}{[$-$0.32, +0.11]} & \ablmaincell{$-$0.22}{[$-$0.50, +0.01]} & \ablmaincell{+0.02}{[$-$0.07, +0.12]} & --- & --- \\
Flight & 1 & \ablmaincell{+0.06}{[$-$0.16, +0.33]} & \ablmaincell{+0.02}{[$-$0.35, +0.45]} & \ablmaincell{+0.02}{[$-$0.35, +0.45]} & \ablmaincell{$-$0.00}{[$-$0.40, +0.46]} & \ablmaincell{+0.02}{[$-$0.31, +0.42]} & \ablmaincell{+0.08}{[$-$0.16, +0.37]} & --- & --- \\
Flight & 2 & \ablmaincell{$-$0.02}{[$-$0.24, +0.27]} & \ablmaincell{+0.05}{[$-$0.23, +0.56]} & \ablmaincell{+0.01}{[$-$0.33, +0.57]} & \ablmaincell{$-$0.07}{[$-$0.49, +0.50]} & \ablmaincell{$-$0.12}{[$-$0.50, +0.43]} & \ablmaincell{+0.03}{[$-$0.12, +0.25]} & --- & --- \\
Flight & 3 & \ablmaincell{$-$0.18}{[$-$0.39, +0.03]} & \ablmaincell{$-$0.11}{[$-$0.37, +0.21]} & \ablmaincell{+0.00}{[$-$0.32, +0.37]} & \ablmaincell{$-$0.13}{[$-$0.45, +0.23]} & \ablmaincell{$-$0.13}{[$-$0.46, +0.22]} & \ablmaincell{$-$0.05}{[$-$0.19, +0.10]} & --- & --- \\
Meeting & 1 & \ablmaincell{$-$0.03}{[$-$0.16, +0.06]} & \ablmaincell{$-$0.05}{[$-$0.19, +0.07]} & \ablmaincell{$-$0.05}{[$-$0.19, +0.07]} & \ablmaincell{$-$0.07$^{\ast}$}{[$-$0.18, $-$0.01]} & \ablmaincell{$-$0.05}{[$-$0.17, +0.02]} & \ablmaincell{$-$0.01}{[$-$0.10, +0.09]} & \ablmaincell{+0.01}{[$-$0.05, +0.09]} & \ablmaincell{+0.02}{[$-$0.01, +0.08]} \\
Meeting & 2 & \ablmaincell{$-$0.05}{[$-$0.15, +0.04]} & \ablmaincell{$-$0.05}{[$-$0.13, +0.03]} & \ablmaincell{$-$0.00}{[$-$0.13, +0.13]} & \ablmaincell{$-$0.10}{[$-$0.23, +0.02]} & \ablmaincell{$-$0.14$^{\ast}$}{[$-$0.27, $-$0.00]} & \ablmaincell{$-$0.03}{[$-$0.07, +0.01]} & \ablmaincell{+0.00}{[$-$0.04, +0.03]} & \ablmaincell{+0.01}{[$-$0.02, +0.05]} \\
Meeting & 4 & \ablmaincell{$-$0.16$^{\ast}$}{[$-$0.29, $-$0.05]} & \ablmaincell{$-$0.17$^{\ast}$}{[$-$0.30, $-$0.08]} & \ablmaincell{$-$0.08}{[$-$0.21, +0.03]} & \ablmaincell{$-$0.17$^{\ast}$}{[$-$0.27, $-$0.07]} & \ablmaincell{$-$0.17$^{\ast}$}{[$-$0.28, $-$0.09]} & \ablmaincell{$-$0.06}{[$-$0.17, +0.02]} & \ablmaincell{+0.00}{[$-$0.03, +0.03]} & \ablmaincell{$-$0.04}{[$-$0.10, +0.01]} \\
\midrule
Animals (Seq.) & 2 & \ablmaincell{+0.05}{[+0.00, +0.10]} & --- & --- & --- & \ablmaincell{+0.03}{[$-$0.01, +0.07]} & --- & --- & --- \\
Animals (Seq.) & 3 & \ablmaincell{+0.02}{[$-$0.01, +0.06]} & --- & --- & --- & \ablmaincell{+0.01}{[$-$0.02, +0.05]} & --- & --- & --- \\
TB-D (Seq.) & 1 & \ablmaincell{+0.00}{[$-$0.01, +0.01]} & --- & --- & --- & \ablmaincell{$-$0.00}{[$-$0.01, +0.01]} & --- & --- & --- \\
TB-D (Seq.) & 2 & \ablmaincell{+0.01}{[$-$0.01, +0.02]} & --- & --- & --- & \ablmaincell{$-$0.00}{[$-$0.01, +0.01]} & --- & --- & --- \\
TB-H & 2 & \ablmaincell{$-$0.04$^{\ast}$}{[$-$0.07, $-$0.01]} & --- & --- & --- & \ablmaincell{$-$0.02$^{\ast}$}{[$-$0.04, $-$0.00]} & --- & --- & --- \\
\midrule
\addlinespace[2pt]
\multicolumn{10}{c}{\textbf{GPT-6.1~Sol}} \\
& & \multicolumn{5}{c}{\textit{Explicit decomposition}} & \multicolumn{3}{c}{\textit{Local-value auxiliaries}} \\
\cmidrule(lr){3-7}\cmidrule(lr){8-10}
Bench. & $B$ & R & PR & S2 & Joint add. & Joint v+M & D+v & Aux.\ loss & Wider \\
\midrule
Animals & 1 & \ablmaincell{+0.03}{[$-$0.02, +0.09]} & \ablmaincell{+0.09}{[$-$0.04, +0.27]} & \ablmaincell{+0.09}{[$-$0.04, +0.27]} & \ablmaincell{$-$0.07}{[$-$0.23, +0.05]} & \ablmaincell{+0.04}{[$-$0.01, +0.11]} & \ablmaincell{+0.01}{[$-$0.00, +0.04]} & --- & --- \\
Animals & 2 & \ablmaincell{+0.01}{[$-$0.15, +0.16]} & \ablmaincell{+0.05}{[$-$0.07, +0.17]} & \ablmaincell{+0.01}{[$-$0.14, +0.16]} & \ablmaincell{+0.03}{[$-$0.13, +0.19]} & \ablmaincell{+0.01}{[$-$0.15, +0.17]} & \ablmaincell{+0.08}{[$-$0.00, +0.18]} & --- & --- \\
Animals & 3 & \ablmaincell{$-$0.06$^{\ast}$}{[$-$0.13, $-$0.00]} & \ablmaincell{$-$0.02}{[$-$0.08, +0.04]} & \ablmaincell{$-$0.13$^{\ast}$}{[$-$0.23, $-$0.04]} & \ablmaincell{$-$0.12}{[$-$0.28, +0.02]} & \ablmaincell{$-$0.08}{[$-$0.23, +0.06]} & \ablmaincell{$-$0.01}{[$-$0.05, +0.02]} & --- & --- \\
Trip & 1 & \ablmaincell{+0.09$^{\ast}$}{[+0.01, +0.21]} & \ablmaincell{+0.10$^{\ast}$}{[+0.01, +0.24]} & \ablmaincell{+0.10$^{\ast}$}{[+0.01, +0.24]} & \ablmaincell{$-$0.03}{[$-$0.15, +0.09]} & \ablmaincell{$-$0.04}{[$-$0.18, +0.07]} & \ablmaincell{+0.04}{[$-$0.02, +0.11]} & --- & --- \\
Trip & 2 & \ablmaincell{$-$0.04}{[$-$0.15, +0.05]} & \ablmaincell{+0.06}{[$-$0.04, +0.17]} & \ablmaincell{$-$0.00}{[$-$0.18, +0.21]} & \ablmaincell{$-$0.14}{[$-$0.36, +0.04]} & \ablmaincell{$-$0.33$^{\ast}$}{[$-$0.59, $-$0.11]} & \ablmaincell{$-$0.04}{[$-$0.16, +0.05]} & --- & --- \\
Flight & 1 & \ablmaincell{+0.16}{[$-$0.12, +0.43]} & \ablmaincell{+0.07}{[$-$0.42, +0.65]} & \ablmaincell{+0.07}{[$-$0.42, +0.65]} & \ablmaincell{+0.00}{[$-$0.45, +0.50]} & \ablmaincell{$-$0.31}{[$-$0.75, +0.08]} & \ablmaincell{$-$0.04}{[$-$0.33, +0.35]} & --- & --- \\
Flight & 2 & \ablmaincell{+0.24}{[$-$0.02, +0.52]} & \ablmaincell{+0.11}{[$-$0.29, +0.50]} & \ablmaincell{+0.36}{[$-$0.27, +1.00]} & \ablmaincell{+0.46$^{\ast}$}{[+0.01, +1.04]} & \ablmaincell{$-$0.09}{[$-$0.57, +0.41]} & \ablmaincell{+0.13}{[$-$0.15, +0.42]} & --- & --- \\
Flight & 3 & \ablmaincell{+0.36}{[$-$0.05, +0.85]} & \ablmaincell{+0.12}{[$-$0.35, +0.66]} & \ablmaincell{+0.47}{[$-$0.19, +1.32]} & \ablmaincell{+0.37}{[$-$0.14, +1.05]} & \ablmaincell{+0.06}{[$-$0.55, +0.72]} & \ablmaincell{+0.10}{[$-$0.36, +0.55]} & --- & --- \\
Meeting & 1 & \ablmaincell{+0.03}{[$-$0.04, +0.13]} & \ablmaincell{+0.06}{[$-$0.07, +0.23]} & \ablmaincell{+0.06}{[$-$0.07, +0.23]} & \ablmaincell{+0.06}{[$-$0.11, +0.25]} & \ablmaincell{+0.06}{[$-$0.15, +0.30]} & \ablmaincell{+0.09}{[$-$0.02, +0.25]} & \ablmaincell{+0.01}{[+0.00, +0.02]} & \ablmaincell{$-$0.03}{[$-$0.09, +0.01]} \\
Meeting & 2 & \ablmaincell{$-$0.08$^{\ast}$}{[$-$0.14, $-$0.01]} & \ablmaincell{$-$0.08}{[$-$0.19, +0.02]} & \ablmaincell{+0.05}{[$-$0.02, +0.14]} & \ablmaincell{+0.03}{[$-$0.11, +0.17]} & \ablmaincell{+0.02}{[$-$0.09, +0.14]} & \ablmaincell{$-$0.05}{[$-$0.14, +0.04]} & \ablmaincell{$-$0.01}{[$-$0.03, +0.00]} & \ablmaincell{$-$0.02}{[$-$0.06, +0.00]} \\
Meeting & 4 & \ablmaincell{$-$0.17$^{\ast}$}{[$-$0.26, $-$0.08]} & \ablmaincell{$-$0.16$^{\ast}$}{[$-$0.25, $-$0.06]} & \ablmaincell{$-$0.09}{[$-$0.20, +0.01]} & \ablmaincell{$-$0.00}{[$-$0.10, +0.10]} & \ablmaincell{$-$0.00}{[$-$0.11, +0.10]} & \ablmaincell{$-$0.03}{[$-$0.09, +0.03]} & \ablmaincell{+0.00}{[$-$0.01, +0.01]} & \ablmaincell{+0.02}{[$-$0.04, +0.08]} \\
\midrule
Animals (Seq.) & 2 & \ablmaincell{+0.05}{[$-$0.00, +0.10]} & --- & --- & --- & \ablmaincell{$-$0.00}{[$-$0.06, +0.05]} & --- & --- & --- \\
Animals (Seq.) & 3 & \ablmaincell{+0.03}{[$-$0.01, +0.07]} & --- & --- & --- & \ablmaincell{+0.02}{[$-$0.02, +0.06]} & --- & --- & --- \\
TB-D (Seq.) & 1 & \ablmaincell{$-$0.00}{[$-$0.01, +0.01]} & --- & --- & --- & \ablmaincell{$-$0.00}{[$-$0.01, +0.01]} & --- & --- & --- \\
TB-D (Seq.) & 2 & \ablmaincell{$-$0.00}{[$-$0.01, +0.01]} & --- & --- & --- & \ablmaincell{$-$0.00}{[$-$0.02, +0.01]} & --- & --- & --- \\
TB-H & 2 & \ablmaincell{+0.01}{[$-$0.01, +0.02]} & --- & --- & --- & \ablmaincell{+0.01}{[$-$0.01, +0.02]} & --- & --- & --- \\
\bottomrule
\end{tabular}
\caption{\textbf{Full method ablation, part I}: explicit decompositions and
local-value auxiliaries on the frozen test sets. Entries are alternative
minus direct-regression normalized gain, with paired $95\%$ bootstrap intervals.
Both arms use five-model ensembles, averaged over three training runs.
$^{\ast}$ denotes an interval excluding zero before rounding;
--- denotes an unevaluated combination. Seq.\ denotes sequential allocation.}
\label{tab:app_ablation}
\end{table*}

\begin{table*}[p]
\centering
\scriptsize
\setlength{\tabcolsep}{6pt}
\renewcommand{\arraystretch}{1.0}
\begin{tabular}{llccccc}
\toprule
\multicolumn{7}{c}{\textbf{Sonnet~5}} \\
& & \multicolumn{3}{c}{\textit{Within-episode / relative supervision}} & \multicolumn{2}{c}{\textit{Simpler allocators}} \\
\cmidrule(lr){3-5}\cmidrule(lr){6-7}
Bench. & $B$ & Within-ep. & Contrast-D & Hybrid & Local Value & Single model \\
\midrule
Animals & 1 & \ablmaincell{+0.10}{[$-$0.02, +0.28]} & \ablmaincell{+0.04}{[$-$0.04, +0.15]} & \ablmaincell{+0.06}{[$-$0.03, +0.18]} & \ablmaincell{+0.12}{[$-$0.00, +0.30]} & \ablmaincell{+0.00}{[$-$0.05, +0.06]} \\
Animals & 2 & \ablmaincell{+0.01}{[$-$0.12, +0.14]} & \ablmaincell{+0.11$^{\ast}$}{[+0.01, +0.23]} & \ablmaincell{$-$0.04}{[$-$0.12, +0.04]} & \ablmaincell{+0.03}{[$-$0.13, +0.17]} & \ablmaincell{$-$0.06$^{\ast}$}{[$-$0.11, $-$0.01]} \\
Animals & 3 & \ablmaincell{$-$0.37$^{\ast}$}{[$-$0.61, $-$0.19]} & \ablmaincell{+0.17}{[$-$0.00, +0.33]} & \ablmaincell{+0.06}{[$-$0.07, +0.18]} & \ablmaincell{$-$0.33$^{\ast}$}{[$-$0.57, $-$0.15]} & \ablmaincell{$-$0.03}{[$-$0.08, +0.01]} \\
Trip & 1 & \ablmaincell{+0.08}{[$-$0.12, +0.36]} & \ablmaincell{+0.06}{[$-$0.15, +0.34]} & \ablmaincell{$-$0.05}{[$-$0.22, +0.07]} & \ablmaincell{+0.02}{[$-$0.20, +0.25]} & \ablmaincell{$-$0.11$^{\ast}$}{[$-$0.19, $-$0.03]} \\
Trip & 2 & \ablmaincell{$-$0.05}{[$-$0.33, +0.20]} & \ablmaincell{+0.01}{[$-$0.25, +0.27]} & \ablmaincell{+0.01}{[$-$0.05, +0.07]} & \ablmaincell{$-$0.04}{[$-$0.25, +0.17]} & \ablmaincell{$-$0.00}{[$-$0.07, +0.08]} \\
Flight & 1 & \ablmaincell{+0.15}{[$-$0.17, +0.53]} & \ablmaincell{$-$0.14}{[$-$0.49, +0.27]} & \ablmaincell{+0.04}{[$-$0.05, +0.16]} & \ablmaincell{$-$0.02}{[$-$0.40, +0.40]} & \ablmaincell{+0.03}{[$-$0.07, +0.20]} \\
Flight & 2 & \ablmaincell{+0.08}{[$-$0.31, +0.74]} & \ablmaincell{$-$0.25}{[$-$0.65, +0.30]} & \ablmaincell{+0.07}{[$-$0.08, +0.27]} & \ablmaincell{$-$0.08}{[$-$0.47, +0.50]} & \ablmaincell{$-$0.02}{[$-$0.19, +0.25]} \\
Flight & 3 & \ablmaincell{+0.02}{[$-$0.42, +0.55]} & \ablmaincell{$-$0.25}{[$-$0.60, +0.16]} & \ablmaincell{$-$0.07}{[$-$0.25, +0.10]} & \ablmaincell{$-$0.37}{[$-$0.74, +0.07]} & \ablmaincell{$-$0.11}{[$-$0.21, +0.01]} \\
Meeting & 1 & \ablmaincell{$-$0.05}{[$-$0.19, +0.06]} & \ablmaincell{$-$0.01}{[$-$0.12, +0.09]} & \ablmaincell{$-$0.00}{[$-$0.06, +0.06]} & \ablmaincell{$-$0.07}{[$-$0.22, +0.04]} & \ablmaincell{$-$0.05}{[$-$0.14, +0.03]} \\
Meeting & 2 & \ablmaincell{$-$0.04}{[$-$0.19, +0.12]} & \ablmaincell{+0.04}{[$-$0.10, +0.20]} & \ablmaincell{+0.03}{[$-$0.03, +0.08]} & \ablmaincell{$-$0.00}{[$-$0.12, +0.13]} & \ablmaincell{+0.00}{[$-$0.03, +0.04]} \\
Meeting & 4 & \ablmaincell{$-$0.08}{[$-$0.17, +0.02]} & \ablmaincell{$-$0.06}{[$-$0.15, +0.01]} & \ablmaincell{$-$0.00}{[$-$0.05, +0.04]} & \ablmaincell{$-$0.09}{[$-$0.19, +0.00]} & \ablmaincell{$-$0.05$^{\ast}$}{[$-$0.10, $-$0.00]} \\
\midrule
Animals (Seq.) & 2 & --- & --- & --- & \ablmaincell{+0.07$^{\ast}$}{[+0.02, +0.12]} & \ablmaincell{$-$0.06$^{\ast}$}{[$-$0.10, $-$0.02]} \\
Animals (Seq.) & 3 & --- & --- & --- & \ablmaincell{+0.02}{[$-$0.02, +0.06]} & \ablmaincell{$-$0.05$^{\ast}$}{[$-$0.08, $-$0.01]} \\
TB-D (Seq.) & 1 & --- & --- & --- & \ablmaincell{$-$0.07$^{\ast}$}{[$-$0.13, $-$0.01]} & \ablmaincell{$-$0.00}{[$-$0.02, +0.02]} \\
TB-D (Seq.) & 2 & --- & --- & --- & \ablmaincell{$-$0.08$^{\ast}$}{[$-$0.12, $-$0.04]} & \ablmaincell{$-$0.01}{[$-$0.02, +0.01]} \\
TB-H & 2 & --- & --- & --- & \ablmaincell{$-$0.11$^{\ast}$}{[$-$0.18, $-$0.05]} & \ablmaincell{$-$0.03$^{\ast}$}{[$-$0.06, $-$0.00]} \\
\midrule
\addlinespace[2pt]
\multicolumn{7}{c}{\textbf{GPT-6.1~Sol}} \\
& & \multicolumn{3}{c}{\textit{Within-episode / relative supervision}} & \multicolumn{2}{c}{\textit{Simpler allocators}} \\
\cmidrule(lr){3-5}\cmidrule(lr){6-7}
Bench. & $B$ & Within-ep. & Contrast-D & Hybrid & Local Value & Single model \\
\midrule
Animals & 1 & \ablmaincell{+0.10}{[$-$0.03, +0.28]} & \ablmaincell{$-$0.06}{[$-$0.21, +0.03]} & \ablmaincell{$-$0.02}{[$-$0.08, +0.02]} & \ablmaincell{+0.09}{[$-$0.04, +0.27]} & \ablmaincell{$-$0.02}{[$-$0.06, +0.03]} \\
Animals & 2 & \ablmaincell{+0.11}{[$-$0.06, +0.27]} & \ablmaincell{+0.09}{[$-$0.05, +0.22]} & \ablmaincell{+0.04}{[$-$0.06, +0.15]} & \ablmaincell{+0.11}{[$-$0.06, +0.27]} & \ablmaincell{+0.00}{[$-$0.06, +0.07]} \\
Animals & 3 & \ablmaincell{$-$0.28$^{\ast}$}{[$-$0.45, $-$0.12]} & \ablmaincell{$-$0.08}{[$-$0.23, +0.07]} & \ablmaincell{$-$0.01}{[$-$0.05, +0.02]} & \ablmaincell{$-$0.31$^{\ast}$}{[$-$0.46, $-$0.16]} & \ablmaincell{$-$0.01}{[$-$0.04, +0.03]} \\
Trip & 1 & \ablmaincell{+0.13$^{\ast}$}{[+0.04, +0.26]} & \ablmaincell{+0.00}{[$-$0.13, +0.14]} & \ablmaincell{+0.03$^{\ast}$}{[+0.01, +0.07]} & \ablmaincell{+0.10$^{\ast}$}{[+0.01, +0.24]} & \ablmaincell{$-$0.04$^{\ast}$}{[$-$0.10, $-$0.00]} \\
Trip & 2 & \ablmaincell{$-$0.18}{[$-$0.44, +0.06]} & \ablmaincell{$-$0.01}{[$-$0.16, +0.15]} & \ablmaincell{$-$0.01}{[$-$0.08, +0.07]} & \ablmaincell{$-$0.05}{[$-$0.27, +0.19]} & \ablmaincell{$-$0.02}{[$-$0.08, +0.04]} \\
Flight & 1 & \ablmaincell{+0.55$^{\ast}$}{[+0.19, +1.05]} & \ablmaincell{$-$0.10}{[$-$0.63, +0.29]} & \ablmaincell{$-$0.03}{[$-$0.24, +0.15]} & \ablmaincell{+0.01}{[$-$0.48, +0.62]} & \ablmaincell{$-$0.02}{[$-$0.14, +0.12]} \\
Flight & 2 & \ablmaincell{+0.12}{[$-$0.35, +0.65]} & \ablmaincell{$-$0.15}{[$-$0.43, +0.14]} & \ablmaincell{$-$0.08}{[$-$0.34, +0.12]} & \ablmaincell{+0.09}{[$-$0.51, +0.80]} & \ablmaincell{+0.03}{[$-$0.09, +0.17]} \\
Flight & 3 & \ablmaincell{+0.69$^{\ast}$}{[+0.04, +1.75]} & \ablmaincell{$-$0.18}{[$-$0.53, +0.19]} & \ablmaincell{$-$0.08}{[$-$0.40, +0.15]} & \ablmaincell{+0.37}{[$-$0.19, +1.15]} & \ablmaincell{+0.01}{[$-$0.16, +0.24]} \\
Meeting & 1 & \ablmaincell{+0.06}{[$-$0.08, +0.22]} & \ablmaincell{+0.04}{[$-$0.08, +0.20]} & \ablmaincell{+0.02}{[$-$0.06, +0.12]} & \ablmaincell{+0.07}{[$-$0.07, +0.24]} & \ablmaincell{$-$0.01}{[$-$0.05, +0.03]} \\
Meeting & 2 & \ablmaincell{+0.10$^{\ast}$}{[+0.02, +0.20]} & \ablmaincell{+0.04}{[$-$0.03, +0.14]} & \ablmaincell{+0.03}{[$-$0.03, +0.12]} & \ablmaincell{+0.06}{[$-$0.03, +0.16]} & \ablmaincell{$-$0.01}{[$-$0.04, +0.02]} \\
Meeting & 4 & \ablmaincell{+0.11$^{\ast}$}{[+0.03, +0.18]} & \ablmaincell{+0.10$^{\ast}$}{[+0.02, +0.18]} & \ablmaincell{+0.06$^{\ast}$}{[+0.01, +0.11]} & \ablmaincell{+0.06}{[$-$0.05, +0.16]} & \ablmaincell{$-$0.02}{[$-$0.05, +0.02]} \\
\midrule
Animals (Seq.) & 2 & --- & --- & --- & \ablmaincell{+0.04}{[$-$0.02, +0.09]} & \ablmaincell{$-$0.09$^{\ast}$}{[$-$0.14, $-$0.04]} \\
Animals (Seq.) & 3 & --- & --- & --- & \ablmaincell{+0.05}{[$-$0.01, +0.12]} & \ablmaincell{$-$0.08$^{\ast}$}{[$-$0.12, $-$0.04]} \\
TB-D (Seq.) & 1 & --- & --- & --- & \ablmaincell{$-$0.07$^{\ast}$}{[$-$0.13, $-$0.01]} & \ablmaincell{$-$0.00}{[$-$0.02, +0.01]} \\
TB-D (Seq.) & 2 & --- & --- & --- & \ablmaincell{$-$0.09$^{\ast}$}{[$-$0.14, $-$0.05]} & \ablmaincell{$-$0.00}{[$-$0.01, +0.01]} \\
TB-H & 2 & --- & --- & --- & \ablmaincell{$-$0.11$^{\ast}$}{[$-$0.18, $-$0.05]} & \ablmaincell{$-$0.01}{[$-$0.03, +0.02]} \\
\bottomrule
\end{tabular}
\caption{\textbf{Full method ablation, part II}: relative supervision and
simpler allocators on the frozen test sets. Format follows
Table~\ref{tab:app_ablation}. Within-ep.\ denotes the within-episode local learner;
Contrast-D uses pairwise ranking loss; Hybrid combines absolute and relative loss.
Local Value uses batch top-$B$ or sequential myopic allocation.
Single model averages individual-model utilities without ensembling;
all other alternatives use five-model ensembles over three training runs.}
\label{tab:app_ablation_b}
\end{table*}

\paragraph{Results across executors.}
Alternatives significantly improve on direct regression in 3 comparisons
and underperform it in 22 under Sonnet~5; the corresponding counts are
13 and 14 under GPT-6.1~Sol. Sol's improvements comprise six alternatives
on Trip at $B=1$, three comparisons on Flight, and four on Meeting.
Residual decompositions R and PR still lose significantly on Meeting
at $B=4$ under both executors, and Local Value loses on batch Animals
at $B=3$ and sequential ThinkingBox. These patterns support direct
regression as a strong default, with task- and executor-dependent
benefits from additional structure.

\paragraph{Additional diagnostics.}
Relative supervision is sensitive to near-tied sets: the two independent label halves agree on only 82--86\% of orderings among top-ranked subset pairs, and the large development gains of Contrast-D did not persist on the frozen test set. We additionally tested uncertainty-aware inputs, uncertainty-weighted local values, and per-task routing on the development split. None improved allocation beyond standard ensembling, and so they were not evaluated on the test set.

\subsection{Controls for direct value regression}
\label{appendix:matched_additive_control}

\paragraph{Matched additive control.}
To separate the effect of set-level supervision from the ability to model
interactions, we construct a joint additive predictor using the direct value
predictor's candidate encoder followed by a local head:
\[
v_i=g_v(\mathrm{Enc}(z_i),c_x),
\qquad
F_{\mathrm{add}}(S)=\sum_{i\in S} v_i .
\]
Thus $F_{\mathrm{add}}(\emptyset)=0$ and the model cannot represent
complementarity or redundancy among consultations. It is trained on the same
set-level targets
$\widehat U_{\pi,L}(S)-\widehat U_{\pi,L}(\emptyset)$
as the direct value predictor, for all labeled sets with
$1\le |S|\le B_{\mathrm{train}}$, where
$B_{\mathrm{train}}=3,2,3,4$ for Animals, Trip, Flight, and Meeting,
respectively. Each set size receives equal total weight, with singleton
examples weighted twice; these choices were fixed on development data.
Inputs are standardized, and the optimizer and training schedule match the
direct predictor.

Table~\ref{tab:app_joint_add} decomposes the gap between a singleton-trained
additive model and the default direct value predictor one factor at a time.
Column~(1) introduces set-level supervision while retaining additivity.
Column~(2) then removes the additive constraint while holding labels,
weighting, and input scaling fixed. Column~(3) switches from standardized to
the raw inputs used by the default direct predictor, and column~(4) restores
the default use of all labeled sets with uniform weighting.

\paragraph{Matched control results.}
Under Sonnet~5, set-level supervision alone significantly improves the additive model on
Meeting at $B=1,4$ (+0.36/+0.30 NG) and Flight at $B=3$ (+0.19).
With supervision and preprocessing matched, relaxing the additive constraint
provides additional significant gains on Animals at $B=3$ (+0.25) and
Meeting at $B=2$ (+0.12), while differences are not significant elsewhere.
Input scaling and the remaining default-training choices also affect some
cells, underscoring the value of the matched comparison in column~(2).

\begin{table}[t]
\centering
\scriptsize
\setlength{\tabcolsep}{3pt}
\renewcommand{\arraystretch}{1.2}
\resizebox{\columnwidth}{!}{%
\begin{tabular}{llcccc}
\toprule
Bench. & $B$ & \makecell{(1) Set vs.\\singleton\\labels} & \makecell{(2) Remove\\additivity} & \makecell{(3) Raw vs.\\standardized\\inputs} & \makecell{(4) All sets,\\uniform\\weights} \\
\midrule
Animals & 1 & \ablcell{+0.01}{[$-$0.05, +0.07]} & \ablcell{+0.01}{[$-$0.05, +0.08]} & \ablcell{$-$0.03}{[$-$0.10, +0.04]} & \ablcell{+0.04}{[$-$0.16, +0.25]} \\
Animals & 2 & \ablcell{+0.14}{[$-$0.08, +0.35]} & \ablcell{+0.06}{[$-$0.02, +0.16]} & \ablcell{$-$0.05}{[$-$0.14, +0.04]} & \ablcell{$-$0.06}{[$-$0.15, +0.01]} \\
Animals & 3 & \ablcell{+0.13}{[$-$0.04, +0.34]} & \ablcell{+0.25$^{\ast}$}{[+0.06, +0.45]} & \ablcell{$-$0.22$^{\ast}$}{[$-$0.39, $-$0.05]} & \ablcell{+0.10$^{\ast}$}{[+0.02, +0.19]} \\
Trip & 1 & \ablcell{+0.10}{[$-$0.08, +0.36]} & \ablcell{$-$0.08}{[$-$0.34, +0.11]} & \ablcell{+0.01}{[$-$0.20, +0.25]} & \ablcell{$-$0.03}{[$-$0.30, +0.20]} \\
Trip & 2 & \ablcell{$-$0.01}{[$-$0.28, +0.20]} & \ablcell{$-$0.15}{[$-$0.38, +0.05]} & \ablcell{+0.21}{[$-$0.03, +0.48]} & \ablcell{$-$0.04}{[$-$0.23, +0.12]} \\
Flight & 1 & \ablcell{+0.12}{[$-$0.22, +0.45]} & \ablcell{+0.02}{[$-$0.32, +0.35]} & \ablcell{+0.22}{[$-$0.23, +0.66]} & \ablcell{+0.04}{[$-$0.13, +0.30]} \\
Flight & 2 & \ablcell{$-$0.02}{[$-$0.30, +0.26]} & \ablcell{+0.06}{[$-$0.34, +0.46]} & \ablcell{+0.21}{[$-$0.35, +0.62]} & \ablcell{+0.08}{[$-$0.31, +0.56]} \\
Flight & 3 & \ablcell{+0.19$^{\ast}$}{[+0.02, +0.42]} & \ablcell{$-$0.17}{[$-$0.48, +0.11]} & \ablcell{+0.34}{[$-$0.12, +0.76]} & \ablcell{+0.15}{[$-$0.13, +0.50]} \\
Meeting & 1 & \ablcell{+0.36$^{\ast}$}{[+0.09, +0.60]} & \ablcell{$-$0.07}{[$-$0.22, +0.09]} & \ablcell{+0.13}{[$-$0.02, +0.31]} & \ablcell{+0.09$^{\ast}$}{[+0.01, +0.20]} \\
Meeting & 2 & \ablcell{+0.12}{[$-$0.10, +0.31]} & \ablcell{+0.12$^{\ast}$}{[+0.01, +0.24]} & \ablcell{+0.09}{[$-$0.05, +0.24]} & \ablcell{+0.01}{[$-$0.04, +0.05]} \\
Meeting & 4 & \ablcell{+0.30$^{\ast}$}{[+0.15, +0.44]} & \ablcell{+0.04}{[$-$0.05, +0.14]} & \ablcell{+0.14$^{\ast}$}{[+0.04, +0.25]} & \ablcell{+0.03}{[$-$0.03, +0.11]} \\
\bottomrule
\end{tabular}}
\caption{
Matched additive interaction control (Sonnet~5, frozen test).
Entries are differences in normalized gain with paired $95\%$ bootstrap
intervals; $^{\ast}$ indicates that the interval excludes zero.
Each column changes one factor along the path from an additive model trained
on singleton labels to the default direct value predictor, so the four
columns sum to the total difference.
(1) Add set-level supervision while retaining additivity.
(2) Remove the additive constraint with labels, weighting, and standardized
inputs matched.
(3) Replace standardized with raw inputs.
(4) Restore the default use of all labeled sets with uniform weighting.
Batch rows use three-model ensembles; Meeting reports single models averaged
over three seeds.
}
\label{tab:app_joint_add}
\end{table}

\paragraph{Matched control under GPT-6.1~Sol.}
Using the same control protocol, removing additivity yields positive
Meeting effects at every budget ($+0.07$ to $+0.15$ NG), significant
at $B=1$; the Animals $B=3$ effect does not replicate ($-0.02$, not
significant). Set-level supervision improves the additive model at
$B=4$ on Meeting ($+0.18$) and at $B=3$ on Animals ($+0.14$), both
significantly. On Flight, removing additivity with standardized inputs
reduces NG by $0.92$ and $0.97$ at $B=2,3$, respectively, while switching
to raw inputs recovers $0.44$ and $0.65$ (all significant).
Thus the effects of additivity and preprocessing depend on the executor.
This matched control retains its separate three-seed protocol:
three-model ensembles for Animals, Trip, and Flight, and single-model
utilities averaged over three seeds for Meeting.

\paragraph{Singleton-value diagnostic.}
We test whether inaccurate singleton estimates impair set-value learning
by replacing the learned local values in D+v, R, PR, and S2 with
$\widehat U_{\pi,L}(\{i\})-\widehat U_{\pi,L}(\emptyset)$
at both training and prediction time.
We use Sonnet~5 development data for Animals, Trip Planning, and Flight Recommendation, with grouped five-fold out-of-fold evaluation and three-model ensembles. Seeds, folds, and hyperparameters are matched between conditions. Models train on the label half, while allocations are scored on the independent evaluation half. The substituted values are privileged estimates: they remove prediction error relative to the stored singleton labels, but retain Monte Carlo noise. Meeting Planning is excluded because its development data lack
independent evaluation-half labels.

For $B\ge2$, the substitution significantly improves all four variants
on Flight by $0.51$--$0.84$ NG. Changes are smaller and mixed on Trip
at $B=2$ ($-0.03$ to $+0.13$), while none is significant on Animals
($-0.07$ to $+0.03$).
Significance is assessed using paired $95\%$ bootstrap intervals over episodes (4,000 resamples). We emphasize $B\ge2$ because the reference oracle at $B=1$ already selects the best label-half singleton. These results support a setting-dependent role for singleton estimation error.
This development diagnostic does not establish the cause of specific
test-set losses.

\section{Efficiency Ablation Details}
\label{appendix:scalability}

Tables~\ref{tab:app_mc}, \ref{tab:app_subset} and \ref{tab:app_search} give the full results underlying Figure~\ref{fig:efficiency}. 
The first two vary, respectively, the number of Monte Carlo rollouts per training label and the number of labeled consultation sets per task, reporting retention and differences from default, with paired $95\%$ confidence intervals for the differences. The third evaluates approximate deployment-time search, reporting agreement with exact subset search, value-predictor evaluations, and normalized gain. Full results are shown for every evaluated benchmark and budget.

\paragraph{Protocol.}
The efficiency ablations use the first main training run's five-model ensemble
(initialization seeds 0--4) as the default reference, rather than the
three-run mean reported in the main results. Each reduced supervision
condition is retrained from scratch with the same training tasks, features,
architecture, optimizer, training schedule, and initialization seeds,
changing only the ablated supervision. Target normalization is recomputed
from each condition's training targets as in the default pipeline. Test tasks,
evaluation labels, and random and oracle references are fixed across
conditions. For Meeting Planning, the default condition labels 43 of the
386 feasible consultation sets per training task.
Consequently, default reference values in this appendix may differ slightly
from the three-run averages in the main tables.

\paragraph{Predictor reuse across budgets.}
In the batch setting, consultation value is independent of the deployment budget $B$, which only restricts the feasible consultation sets. Our DeepSets predictor is cardinality-agnostic and can therefore be reused when $B$ changes. Prior work has demonstrated DeepSets models evaluated on substantially larger sets than those observed during training \citep{zaheer2017deepsets}. Thus, increasing the budget does not require a new model or retraining from scratch. If the new budget introduces set sizes absent from training, however, their values are extrapolations, and additional larger-set supervision may be useful for fine-tuning. In the sequential setting, continuation value depends explicitly on the remaining budget $b_t$, so the predictor is trained with $b_t$ as an input and is most naturally reused within the budget range covered during training.

\begin{table*}[t]
\centering
\scriptsize
\setlength{\tabcolsep}{3pt}
\begin{tabular}{lllrrcr}
\toprule
Bench. & $B$ & Condition & $R_{\rm train}$ by $|S|$ & Ensemble result & $\Delta$ vs.\ default [95\% CI] & Retention \\
\midrule
Animals & 1 & R1 & (1, 1, 1, 1) & 0.549 & -0.005\,{\scriptsize[-0.056,+0.042]} & 0.99 \\
Animals & 1 & R2 & (2, 2, 2, 1) & 0.554 & +0.000\,{\scriptsize[+0.000,+0.000]} & 1.00 \\
Animals & 1 & R4 & (4, 3, 2, 1) & 0.650 & +0.096\,{\scriptsize[+0.000,+0.289]} & 1.17 \\
Animals & 1 & default & (8, 3, 2, 1) & 0.554 & +0.000\,{\scriptsize[+0.000,+0.000]} & 1.00 \\
Animals & 2 & R1 & (1, 1, 1, 1) & 0.606 & +0.022\,{\scriptsize[-0.026,+0.093]} & 1.04 \\
Animals & 2 & R2 & (2, 2, 2, 1) & 0.558 & -0.026\,{\scriptsize[-0.064,+0.000]} & 0.96 \\
Animals & 2 & R4 & (4, 3, 2, 1) & 0.584 & -0.000\,{\scriptsize[-0.026,+0.022]} & 1.00 \\
Animals & 2 & default & (8, 3, 2, 1) & 0.584 & +0.000\,{\scriptsize[+0.000,+0.000]} & 1.00 \\
Animals & 3 & R1 & (1, 1, 1, 1) & 0.597 & -0.007\,{\scriptsize[-0.018,+0.000]} & 0.99 \\
Animals & 3 & R2 & (2, 2, 2, 1) & 0.593 & -0.011\,{\scriptsize[-0.057,+0.029]} & 0.98 \\
Animals & 3 & R4 & (4, 3, 2, 1) & 0.604 & +0.000\,{\scriptsize[+0.000,+0.000]} & 1.00 \\
Animals & 3 & default & (8, 3, 2, 1) & 0.604 & +0.000\,{\scriptsize[+0.000,+0.000]} & 1.00 \\
Flight & 1 & R1 & (1, 1) & 0.485 & +0.094\,{\scriptsize[-0.065,+0.311]} & 1.24 \\
Flight & 1 & R2 & (2, 2) & 0.447 & +0.056\,{\scriptsize[-0.062,+0.246]} & 1.14 \\
Flight & 1 & default & (4, 2) & 0.391 & +0.000\,{\scriptsize[+0.000,+0.000]} & 1.00 \\
Flight & 2 & R1 & (1, 1) & 0.685 & +0.187\,{\scriptsize[-0.123,+0.541]} & 1.37 \\
Flight & 2 & R2 & (2, 2) & 0.399 & -0.099\,{\scriptsize[-0.345,+0.199]} & 0.80 \\
Flight & 2 & default & (4, 2) & 0.498 & +0.000\,{\scriptsize[+0.000,+0.000]} & 1.00 \\
Flight & 3 & R1 & (1, 1) & 0.335 & +0.006\,{\scriptsize[-0.270,+0.343]} & 1.02 \\
Flight & 3 & R2 & (2, 2) & 0.281 & -0.049\,{\scriptsize[-0.301,+0.252]} & 0.85 \\
Flight & 3 & default & (4, 2) & 0.330 & +0.000\,{\scriptsize[+0.000,+0.000]} & 1.00 \\
Trip & 1 & R1 & 1 & 0.729 & -0.157\,{\scriptsize[-0.397,+0.036]} & 0.82 \\
Trip & 1 & R2 & 2 & 0.658 & -0.228\,{\scriptsize[-0.466,-0.024]}$^{\ast}$ & 0.74 \\
Trip & 1 & R5 & 5 & 0.915 & +0.029\,{\scriptsize[-0.023,+0.103]} & 1.03 \\
Trip & 1 & default & 10 & 0.886 & +0.000\,{\scriptsize[+0.000,+0.000]} & 1.00 \\
Trip & 2 & R1 & 1 & 0.916 & +0.108\,{\scriptsize[-0.080,+0.313]} & 1.13 \\
Trip & 2 & R2 & 2 & 0.948 & +0.140\,{\scriptsize[+0.008,+0.284]}$^{\ast}$ & 1.17 \\
Trip & 2 & R5 & 5 & 0.894 & +0.086\,{\scriptsize[-0.010,+0.194]} & 1.11 \\
Trip & 2 & default & 10 & 0.808 & +0.000\,{\scriptsize[+0.000,+0.000]} & 1.00 \\
Meeting & 1 & R1/8 & (4, 2, 1, 1, 1) & 2.558 & -0.064\,{\scriptsize[-0.117,-0.020]}$^{\ast}$ & 0.81 \\
Meeting & 1 & R1/4 & (8, 3, 2, 2, 2) & 2.601 & -0.022\,{\scriptsize[-0.074,+0.034]} & 0.93 \\
Meeting & 1 & R1/2 & (16, 6, 4, 4, 4) & 2.594 & -0.029\,{\scriptsize[-0.070,-0.003]}$^{\ast}$ & 0.91 \\
Meeting & 1 & default & (32, 12, 8, 8, 8) & 2.623 & +0.000\,{\scriptsize[+0.000,+0.000]} & 1.00 \\
Meeting & 2 & R1/8 & (4, 2, 1, 1, 1) & 3.000 & +0.011\,{\scriptsize[-0.070,+0.093]} & 1.02 \\
Meeting & 2 & R1/4 & (8, 3, 2, 2, 2) & 3.006 & +0.017\,{\scriptsize[-0.055,+0.093]} & 1.03 \\
Meeting & 2 & R1/2 & (16, 6, 4, 4, 4) & 3.006 & +0.017\,{\scriptsize[-0.050,+0.093]} & 1.03 \\
Meeting & 2 & default & (32, 12, 8, 8, 8) & 2.989 & +0.000\,{\scriptsize[+0.000,+0.000]} & 1.00 \\
Meeting & 4 & R1/8 & (4, 2, 1, 1, 1) & 3.796 & -0.168\,{\scriptsize[-0.333,-0.040]}$^{\ast}$ & 0.86 \\
Meeting & 4 & R1/4 & (8, 3, 2, 2, 2) & 3.953 & -0.011\,{\scriptsize[-0.082,+0.061]} & 0.99 \\
Meeting & 4 & R1/2 & (16, 6, 4, 4, 4) & 3.943 & -0.021\,{\scriptsize[-0.114,+0.071]} & 0.98 \\
Meeting & 4 & default & (32, 12, 8, 8, 8) & 3.964 & +0.000\,{\scriptsize[+0.000,+0.000]} & 1.00 \\
\bottomrule
\end{tabular}
\caption{\textbf{MC supervision efficiency on the frozen test set (Sonnet 5)}. 
We vary the number of MC samples used to estimate each training label.
Because one sample can require multiple executor calls, Figure~\ref{fig:mc_efficiency} reports actual training-call fractions rather than $R_{\rm train}$ fractions. The $R1$ condition uses 62\%, 48\%, and 10\% of default training calls on Animals, Flight, and Trip, respectively; Meeting conditions $R1/8$, $R1/4$, and $R1/2$ use 12.5\%, 25\%, and 50\%. The default condition is the first main training run's five-model ensemble (seeds 0--4); reduced conditions are retrained with the same setup while changing only the available rollout supervision. Performance is normalized gain for Animals, Trip Planning, and Flight Recommendation, and native utility for Meeting Planning. $\Delta$ is measured in the corresponding units. For Trip Planning, reduced conditions subsample the ten systematic hidden-date draws. For Meeting Planning, they subsample the training rollout streams in fixed order.
Animals tuples index set-size groups $\{0,1\}$, $\{2\}$, $\{3\}$, and $\{4,5,6\}$. Flight uses $\{0,1\}$ and
$\{2,3\}$. Meeting lists sizes $0$--$4$ individually. Trip uses a common count across set sizes.
}
\label{tab:app_mc}
\end{table*}

\begin{table*}[t]
\centering
\scriptsize
\setlength{\tabcolsep}{3pt}
\begin{tabular}{lllrrrrcr}
\toprule
Bench. & $B$ & Condition & Labeled sets by $|S|$ & \% of sets & Ensemble result & SD (seeds) & $\Delta$ vs.\ default [95\% CI] & Retention \\
\midrule
Animals & 1 & M7 & (1,6,0,0,0,0,0) & 10.9\% & 0.709 & 0.000 & +0.155\,{\scriptsize[-0.019,+0.421]} & 1.28 \\
Animals & 1 & M16 & (1,6,3,3,2,1,0) & 25.0\% & 0.615 & 0.057 & +0.061\,{\scriptsize[-0.025,+0.212]} & 1.11 \\
Animals & 1 & M32 & (1,6,6,9,6,3,1) & 50.0\% & 0.628 & 0.043 & +0.074\,{\scriptsize[-0.006,+0.224]} & 1.13 \\
Animals & 1 & D42 & (1,6,15,20,0,0,0) & 65.6\% & 0.549 & 0.000 & -0.005\,{\scriptsize[-0.056,+0.042]} & 0.99 \\
Animals & 1 & default & all & 100.0\% & 0.554 & 0.000 & +0.000\,{\scriptsize[+0.000,+0.000]} & 1.00 \\
Animals & 2 & M7 & (1,6,0,0,0,0,0) & 10.9\% & 0.510 & 0.000 & -0.074\,{\scriptsize[-0.256,+0.099]} & 0.87 \\
Animals & 2 & M16 & (1,6,3,3,2,1,0) & 25.0\% & 0.585 & 0.048 & +0.001\,{\scriptsize[-0.057,+0.077]} & 1.00 \\
Animals & 2 & M32 & (1,6,6,9,6,3,1) & 50.0\% & 0.563 & 0.029 & -0.020\,{\scriptsize[-0.079,+0.042]} & 0.97 \\
Animals & 2 & D42 & (1,6,15,20,0,0,0) & 65.6\% & 0.613 & 0.000 & +0.029\,{\scriptsize[-0.029,+0.106]} & 1.05 \\
Animals & 2 & default & all & 100.0\% & 0.584 & 0.000 & +0.000\,{\scriptsize[+0.000,+0.000]} & 1.00 \\
Animals & 3 & M7 & (1,6,0,0,0,0,0) & 10.9\% & 0.332 & 0.000 & -0.272\,{\scriptsize[-0.465,-0.089]}$^{\ast}$ & 0.55 \\
Animals & 3 & M16 & (1,6,3,3,2,1,0) & 25.0\% & 0.575 & 0.036 & -0.029\,{\scriptsize[-0.098,+0.038]} & 0.95 \\
Animals & 3 & M32 & (1,6,6,9,6,3,1) & 50.0\% & 0.613 & 0.045 & +0.010\,{\scriptsize[-0.033,+0.060]} & 1.02 \\
Animals & 3 & D42 & (1,6,15,20,0,0,0) & 65.6\% & 0.568 & 0.000 & -0.036\,{\scriptsize[-0.147,+0.071]} & 0.94 \\
Animals & 3 & default & all & 100.0\% & 0.604 & 0.000 & +0.000\,{\scriptsize[+0.000,+0.000]} & 1.00 \\
Trip & 1 & M4 & (1,3,0,0) & 50.0\% & 0.969 & 0.000 & +0.083\,{\scriptsize[-0.101,+0.252]} & 1.09 \\
Trip & 1 & M5 & (1,3,1,0) & 62.5\% & 0.912 & 0.046 & +0.026\,{\scriptsize[-0.138,+0.191]} & 1.03 \\
Trip & 1 & M7 & (1,3,3,0) & 87.5\% & 1.037 & 0.000 & +0.151\,{\scriptsize[+0.021,+0.303]}$^{\ast}$ & 1.17 \\
Trip & 1 & default & all & 100.0\% & 0.886 & 0.000 & +0.000\,{\scriptsize[+0.000,+0.000]} & 1.00 \\
Trip & 2 & M4 & (1,3,0,0) & 50.0\% & 0.983 & 0.000 & +0.175\,{\scriptsize[+0.003,+0.391]}$^{\ast}$ & 1.22 \\
Trip & 2 & M5 & (1,3,1,0) & 62.5\% & 0.833 & 0.069 & +0.025\,{\scriptsize[-0.128,+0.204]} & 1.03 \\
Trip & 2 & M7 & (1,3,3,0) & 87.5\% & 0.823 & 0.000 & +0.015\,{\scriptsize[-0.126,+0.189]} & 1.02 \\
Trip & 2 & default & all & 100.0\% & 0.808 & 0.000 & +0.000\,{\scriptsize[+0.000,+0.000]} & 1.00 \\
Meeting & 1 & M11 & (1,10,0,0,0) & 2.8\% & 2.613 & 0.000 & -0.010\,{\scriptsize[-0.047,+0.023]} & 0.97 \\
Meeting & 1 & M20 & (1,10,3,2,4) & 5.2\% & 2.599 & 0.020 & -0.023\,{\scriptsize[-0.041,-0.009]}$^{\ast}$ & 0.93 \\
Meeting & 1 & M30 & (1,10,6,5,8) & 7.8\% & 2.613 & 0.001 & -0.010\,{\scriptsize[-0.021,+0.000]} & 0.97 \\
Meeting & 1 & default & (1,10,9,9,14) & 11.1\% & 2.623 & 0.000 & +0.000\,{\scriptsize[+0.000,+0.000]} & 1.00 \\
Meeting & 2 & M11 & (1,10,0,0,0) & 2.8\% & 2.916 & 0.000 & -0.073\,{\scriptsize[-0.182,+0.041]} & 0.87 \\
Meeting & 2 & M20 & (1,10,3,2,4) & 5.2\% & 3.013 & 0.007 & +0.024\,{\scriptsize[-0.006,+0.063]} & 1.04 \\
Meeting & 2 & M30 & (1,10,6,5,8) & 7.8\% & 3.006 & 0.014 & +0.017\,{\scriptsize[-0.012,+0.053]} & 1.03 \\
Meeting & 2 & default & (1,10,9,9,14) & 11.1\% & 2.989 & 0.000 & +0.000\,{\scriptsize[+0.000,+0.000]} & 1.00 \\
Meeting & 4 & M11 & (1,10,0,0,0) & 2.8\% & 3.404 & 0.000 & -0.560\,{\scriptsize[-0.790,-0.341]}$^{\ast}$ & 0.52 \\
Meeting & 4 & M20 & (1,10,3,2,4) & 5.2\% & 3.977 & 0.033 & +0.012\,{\scriptsize[-0.076,+0.107]} & 1.01 \\
Meeting & 4 & M30 & (1,10,6,5,8) & 7.8\% & 3.983 & 0.061 & +0.018\,{\scriptsize[-0.048,+0.093]} & 1.02 \\
Meeting & 4 & default & (1,10,9,9,14) & 11.1\% & 3.964 & 0.000 & +0.000\,{\scriptsize[+0.000,+0.000]} & 1.00 \\
\bottomrule
\end{tabular}
\caption{\textbf{Subset supervision efficiency on the frozen test set (Sonnet~5)}. 
We vary the number of labeled consultation sets per training task while
holding the rollout schedule fixed. Results are averaged over three nested
subset-sampling seeds, with SD across seeds. $\Delta$ and its 95\% confidence
interval are computed from the per-task mean across these seeds using a paired
bootstrap against the default condition. For Animals, D42 labels exactly the
42 sets with $|S|\leq 3$ and provides no supervision for larger sets.
}
\label{tab:app_subset}
\end{table*}

\begin{table*}[ht]
\centering
\scriptsize
\setlength{\tabcolsep}{3pt}
\begin{tabular}{lllcccc}
\toprule
Executor & Bench. & $B$ & Exact & Greedy & Beam(3) & Greedy+swap \\
\midrule
Sonnet 5 & Animals & 1 & 1.00 / 0.55 / 7 & 1.00 / 0.55 / 7 & 1.00 / 0.55 / 7 & 1.00 / 0.55 / 7 \\
Sonnet 5 & Animals & 2 & 1.00 / 0.58 / 22 & 1.00 / 0.58 / 12 & 1.00 / 0.58 / 19 & 1.00 / 0.58 / 16 \\
Sonnet 5 & Animals & 3 & 1.00 / 0.60 / 42 & 0.89 / 0.54 / 16 & 1.00 / 0.60 / 28 & 1.00 / 0.60 / 24 \\
Sonnet 5 & Trip & 1 & 1.00 / 0.89 / 4 & 1.00 / 0.89 / 4 & 1.00 / 0.89 / 4 & 1.00 / 0.89 / 4 \\
Sonnet 5 & Trip & 2 & 1.00 / 0.81 / 7 & 0.87 / 0.77 / 6 & 1.00 / 0.81 / 7 & 1.00 / 0.81 / 7 \\
Sonnet 5 & Flight & 1 & 1.00 / 0.39 / 6 & 1.00 / 0.39 / 6 & 1.00 / 0.39 / 6 & 1.00 / 0.39 / 6 \\
Sonnet 5 & Flight & 2 & 1.00 / 0.50 / 16 & 1.00 / 0.50 / 10 & 1.00 / 0.50 / 15 & 1.00 / 0.50 / 12 \\
Sonnet 5 & Flight & 3 & 1.00 / 0.33 / 26 & 0.97 / 0.31 / 12 & 1.00 / 0.33 / 21 & 0.97 / 0.31 / 14 \\
GPT-6.1 Sol & Animals & 1 & 1.00 / 0.44 / 7 & 1.00 / 0.44 / 7 & 1.00 / 0.44 / 7 & 1.00 / 0.44 / 7 \\
GPT-6.1 Sol & Animals & 2 & 1.00 / 0.34 / 22 & 1.00 / 0.34 / 12 & 1.00 / 0.34 / 19 & 1.00 / 0.34 / 16 \\
GPT-6.1 Sol & Animals & 3 & 1.00 / 0.43 / 42 & 0.67 / 0.41 / 16 & 0.88 / 0.42 / 28 & 1.00 / 0.43 / 25 \\
GPT-6.1 Sol & Trip & 1 & 1.00 / 0.83 / 4 & 1.00 / 0.83 / 4 & 1.00 / 0.83 / 4 & 1.00 / 0.83 / 4 \\
GPT-6.1 Sol & Trip & 2 & 1.00 / 0.73 / 7 & 0.99 / 0.72 / 6 & 1.00 / 0.73 / 7 & 1.00 / 0.73 / 7 \\
GPT-6.1 Sol & Flight & 1 & 1.00 / -0.11 / 6 & 1.00 / -0.11 / 6 & 1.00 / -0.11 / 6 & 1.00 / -0.11 / 6 \\
GPT-6.1 Sol & Flight & 2 & 1.00 / -0.39 / 16 & 0.90 / -0.33 / 10 & 1.00 / -0.39 / 15 & 0.93 / -0.33 / 12 \\
GPT-6.1 Sol & Flight & 3 & 1.00 / -0.57 / 26 & 0.87 / -0.62 / 12 & 1.00 / -0.57 / 21 & 0.90 / -0.61 / 14 \\
Sonnet 5 & Meeting & 1 & 1.00 / 0.97 / 11 & 1.00 / 0.97 / 11 & 1.00 / 0.97 / 11 & 1.00 / 0.97 / 11 \\
Sonnet 5 & Meeting & 2 & 1.00 / 0.82 / 56 & 0.95 / 0.85 / 20 & 1.00 / 0.82 / 35 & 1.00 / 0.82 / 28 \\
Sonnet 5 & Meeting & 4 & 1.00 / 0.98 / 386 & 1.00 / 0.98 / 35 & 1.00 / 0.98 / 75 & 1.00 / 0.98 / 55 \\
GPT-6.1 Sol & Meeting & 1 & 1.00 / 0.72 / 11 & 1.00 / 0.72 / 11 & 1.00 / 0.72 / 11 & 1.00 / 0.72 / 11 \\
GPT-6.1 Sol & Meeting & 2 & 1.00 / 0.78 / 56 & 1.00 / 0.78 / 20 & 1.00 / 0.78 / 35 & 1.00 / 0.78 / 28 \\
GPT-6.1 Sol & Meeting & 4 & 1.00 / 0.84 / 386 & 1.00 / 0.84 / 35 & 1.00 / 0.84 / 75 & 1.00 / 0.84 / 55 \\
\bottomrule
\end{tabular}
\caption{\textbf{Approximate subset search on four batch benchmarks under both executors}. Each cell reports agreement with exact set selected by evaluating all budget-feasible subsets, normalized gain, and the number of value-predictor evaluations per task. Here, normalized gain is reported unclipped. It may exceed 1 because the ex-ante oracle is selected using the label half and evaluated on an independent outcome half, and values below 0 indicate performance worse than random allocation.}
\label{tab:app_search}
\end{table*}

\section{Executor Dependence and Reuse of Supervision Details}
\label{appendix:xfer}

Table~\ref{tab:app_xfer} provides the full diagnostics for the
cross-executor transfer experiment summarized in the main paper.
In addition to transfer performance, we report the correlation between
Sonnet~5 and GPT-6.1~Sol consultation values and how often their
executor-specific predictors induce different allocations.
Transfer is nearly lossless when the executors have similar
consultation-value structure, as on Trip Planning and TB-D, but degrades
when their values and resulting allocations differ substantially, as on
Animals. Meeting Planning also shows that high value correlation does not
guarantee symmetric transfer, while Flight Recommendation illustrates that supervision
from another executor can sometimes outperform supervision from the
target executor itself.

\begin{table*}[ht]
\centering
\small
\setlength{\tabcolsep}{4pt}
\begin{tabular}{lcccccccccc}
\toprule
& & \multicolumn{2}{c}{\textbf{Value correlation}} & & \multicolumn{3}{c}{\textbf{Target: GPT-6.1 Sol}} & \multicolumn{3}{c}{\textbf{Target: Sonnet 5}} \\
\cmidrule(lr){3-4}\cmidrule(lr){6-8}\cmidrule(lr){9-11}
\textbf{Benchmark} & $B$ & raw & disatt. & $P(\mathrm{diff})$ & Own & Sonnet$\rightarrow$ & Prompt & Own & Sol$\rightarrow$ & Prompt \\
\midrule
        Mixed 20Q--Animals & 3 & 0.43 & 0.45 & 0.93 & 0.40 & 0.15$^{\ast}$ & 0.19 & 0.58 & 0.14$^{\ast}$ & 0.16 \\
        NaturalPlan Trip Planning & 2 & 0.93 & 0.97 & 0.22 & 0.80 & 0.85 & 0.31 & 0.83 & 0.84 & 0.09 \\
        NaturalPlan Meeting Planning & 4 & 0.77 & 0.95 & 0.50 & 0.82 & 0.91$^{\ast}$ & 0.22 & 1.00 & 0.89$^{\ast}$ & 0.00 \\
        Flight Recommendation & 3 & 0.43 & 0.46 & 0.87 & 0.00 & 0.25$^{\ast}$ & 0.29 & 0.33 & 0.08 & 0.24 \\
        ThinkingBox-D & 1 & 0.97 & 1.00 & 0.13 & 0.97 & 0.99 & 0.54 & 1.00 & 1.00 & 0.58 \\
\bottomrule
\end{tabular}
\caption{\textbf{Full cross-executor transfer diagnostics} on the frozen test set at the
main result budgets.
\emph{Value correlation} is the correlation between Sonnet~5 and
GPT-6.1~Sol consultation values within each subset size, averaged over
$1 \leq |S|\leq B$; \emph{disatt.} corrects the raw correlation for estimated
label noise using the two executors' label reliabilities.
$P(\mathrm{diff})$ is the fraction of episodes for which the
executor-specific predictors induce different selected sets.
The remaining columns report clipped normalized gain on the target
executor: \emph{Own} uses a predictor trained on that executor's labels,
arrows denote transfer from the other executor, and \emph{Prompt} is
direct LLM allocation.
$^{\ast}$ indicates that transferred and own performance differ
significantly under a paired $95\%$ bootstrap.}
\label{tab:app_xfer}
\end{table*}

\end{document}